\pdfoutput=1
\documentclass[11pt]{article}

\usepackage[final]{acl}

\usepackage{times}
\usepackage{latexsym}

\usepackage[T1]{fontenc}
\usepackage[utf8]{inputenc}

\usepackage{microtype}

\usepackage{inconsolata}

\usepackage{graphicx}

\usepackage{tabularx}
\usepackage{booktabs}
\usepackage{multirow}
\usepackage{graphicx}
\usepackage{amsmath}
\usepackage{cleveref}
\usepackage{xcolor}
 
\usepackage{graphicx}
\usepackage{float} % for [H]
\usepackage{placeins} % for 
\usepackage{caption}
\usepackage{subcaption}
\usepackage{hyperref}

\newcommand*\samethanks[1][\value{footnote}]{\footnotemark[#1]}

\title{The Accuracy Paradox in RLHF: When Better Reward Models Don't Yield Better Language Models}

\author{
  Yanjun Chen$^{1,3}$, Dawei Zhu$^{2}$, Yirong Sun$^{3}$, Xinghao Chen$^{1,3}$, Wei Zhang$^{3}$\thanks{Corresponding authors.}, Xiaoyu Shen$^{3}$\samethanks\\
  $^1$ Department of Computing, The Hong Kong Polytechnic University\\
  $^2$ Saarland University, Saarland Informatics\\
  $^3$ Digital Twin Institute, Eastern Institute of Technology, Ningbo, China \\
  \texttt{yan-jun.chen@connect.polyu.hk \qquad \{zhw,xyshen\}@eitech.edu.cn}
}

\begin{document}
\maketitle
\begin{abstract}
Reinforcement Learning from Human Feedback significantly enhances Natural Language Processing by aligning language models with human expectations. A critical factor in this alignment is the strength of reward models used during training. This study explores whether stronger reward models invariably lead to better language models. 
In this paper, through experiments on relevance, factuality, and completeness tasks using the QA-FEEDBACK dataset and reward models based on Longformer, we uncover a surprising paradox: \emph{language models trained with moderately accurate reward models outperform those guided by highly accurate ones}. This challenges the widely held belief that stronger reward models always lead to better language models, and opens up new avenues for future research into the key factors driving model performance and how to choose the most suitable reward models. Code and additional details are available at \href{https://github.com/EIT-NLP/AccuracyParadox-RLHF}{https://github.com/EIT-NLP/AccuracyParadox-RLHF}.
\end{abstract}

\section{Introduction}
Language models (LMs) have made remarkable progress, achieving close-to-human capabilities in a wide range of tasks~\cite{shen2017estimation,radford2019language,brown2020language,su2022welm,achiam2023gpt,yuan2024self}. While traditional fine-tuning has been effective, it often suffers from exposure bias, where models are trained on ground truth data rather than their own predictions, leading to inconsistencies during generation~\cite{shen2019select,wang2020exposure}. Additionally, fine-tuning lacks the ability to optimize for sequence-level rewards, limiting its effectiveness in capturing complex, human-like preferences~\cite{zhu2024preference}. RLHF addresses these limitations by incorporating feedback from humans, allowing models to generate more contextually relevant and aligned outputs~\cite{stiennon2020learning,ouyang2022training,su2024unravelingmysteryscalinglaws,madaan2024self}. 

It is commonly assumed that higher accuracy in reward models enhances language model performance because these models provide precise feedback during training~\cite{chaudhari2024rlhf}. This perspective suggests that accurate feedback directly improves the effectiveness of LMs, especially in complex tasks like machine translation and question answering~\cite{bai2022training}.

%In this paper, we challenge the common belief that higher accuracy in reward models always leads to improved language model performance, demonstrating that the relationship is not always straightforward. Our findings indicate that overly accurate reward models may not necessarily enhance performance and can actually result in suboptimal outcomes, such as overfitting or poor generalization~\cite{konda1999actor}. 

In this paper, we conducted extensive experiments using the QA-FEEDBACK dataset~\cite{wu2024fine}. Reward models based on Longformer~\cite{beltagy2020longformer} were evaluated for their binary classification accuracy in predicting task relevance, factuality, and completeness. To ensure fair evaluation, the performance of LMs trained with these reward models was assessed using independent high-accuracy models tailored to each task. Surprisingly, our findings reveal a paradox: LMs achieve their best performance \emph{not with the most accurate reward models, but with those of moderate accuracy}~\cite{casper2023open}, challenging the prevailing assumption that higher reward model accuracy directly correlates with improved outcomes. This result raises important questions about the relationship between reward model accuracy and language model performance in RLHF, warranting further investigation.

The main contributions of this study include:
\begin{itemize}
    \item Demonstrating that moderate reward model accuracy and balanced training lead to better language model performance, contradicting the assumption that higher accuracy is invariably beneficial.
    \item Providing insights into reward dynamics, revealing that moderately accurate reward models offer more task-appropriate rewards, which are intuitively more beneficial for training LMs than those provided by the most accurate models.
    \item Analyzing KL divergence trends, showing that moderately accurate reward models facilitate a balanced and stable training process, promoting better generalization and challenging the notion that higher accuracy alone ensures optimal training outcomes.
\end{itemize}

% 动机和问题设置
\section{Motivation and Problem Settings}

\paragraph{Motivation.}

Findings indicate that the strength of reward models in RLHF does not consistently correlate with improved language model performance, challenging the assumption that stronger reward models always lead to better outcomes~\cite{casper2023open}. Understanding the dynamic relationship between reward model accuracy and language model performance is essential for optimizing RLHF in complex NLP tasks~\cite{ouyang2022training}. This study posits that there exists an optimal range of reward model accuracy that maximizes language model performance~\cite{wu2024fine}. Therefore, the primary aim of this research is to identify this optimal range and examine its implications for various NLP applications.

\paragraph{Problem Settings.}

This study investigates the effect of reward model strength on language model performance in RLHF, focusing on tasks that evaluate the factuality, relevance, and completeness of generated text~\cite{wu2024fine}. Specifically, reward model strength is defined by binary classification accuracy on test sets~\cite{wu2024fine}, and language model performance is measured using high-accuracy, independent reward models.

Formally, for a language model trained with RLHF, this study analyzes how the reward model's classification accuracy (\(\mathcal{S}_{\text{RM}}\)) and the number of training steps (\(\tau\)) affect language model performance (\(\mathcal{P}_{\text{LM}}\))~\cite{qin2024towards}. This relationship is mathematically represented by:

\begin{equation}
\mathcal{P}_{\text{LM}} = f(\mathcal{S}_{\text{RM}}, \tau)
\end{equation}

The objective is to determine the optimal conditions that maximize language model performance across various tasks, providing insights for the development of more effective RLHF strategies in NLP~\cite{li2023remax}.

% 实验部分
\section{Experiment and Results}

\subsection{Basic Experimental Setup}

\paragraph{Models.}

We examine three models from the T5 language model family~\cite{raffel2020exploring, kaplan2020scaling}: T5-small\footnote{\url{https://huggingface.co/t5-small}}, T5-base\footnote{\url{https://huggingface.co/t5-base}}, and T5-large\footnote{\url{https://huggingface.co/t5-large}}. Each model underwent supervised fine-tuning (SFT). Reward models were based on Longformer-base-4096, suitable for processing long sequences, necessary for tasks requiring extensive context~\cite{beltagy2020longformer}. These models were trained for tasks involving factuality, relevance, and completeness, with training steps and accuracy ranges summarized in Table \ref{tab:reward_models}.

\begin{table}[h!]
\footnotesize
\centering
\scalebox{1}{%
\begin{tabular}{@{}lcc@{}}
\toprule[1.0pt]
\textbf{Task Type} & \textbf{Steps Range} & \textbf{Accuracies Range} \\ \midrule[1.0pt]
Factuality & 2--1256 & 0.64--0.77 \\ 
Relevance & 2--2852 & 0.49--0.69 \\ 
Completeness & 30--5730 & 0.44--0.70 \\ \bottomrule[1.0pt]
\end{tabular}
}
\caption{\small Training steps and accuracy ranges for reward models by task type.}
\label{tab:reward_models}
\end{table}

\paragraph{Datasets.}

The QA-FEEDBACK dataset~\cite{wu2024fine}, derived from the ASQA dataset~\cite{stelmakh2022asqa}, is used for this study. This dataset focuses on generating long-form answers to ambiguous factual questions in an open-domain setting. The data is split into 3,853/500/948 for training, validation, and testing, requiring the generation of detailed answers from multiple knowledge passages~\cite{Min_Michael_Hajishirzi_Zettlemoyer_2020}.

\paragraph{Hyperparameter Settings.}

We follow the hyperparameter settings recommended by Wu et al.~\cite{wu2024fine}, whose configuration has been specifically designed and empirically validated for RLHF tasks involving QA-feedback. These settings are selected to ensure an optimal trade-off between model performance and training stability, based on prior experimental findings. For a detailed description of all hyperparameters used in the experiments, please refer to Appendix \ref{appendix:hyperparameters}.

\paragraph{Training and Evaluation Paradigm.}

Following common practice~\cite{schulman2017proximal}, we begin by fine-tuning LMs, followed by applying RLHF using Proximal Policy Optimization (PPO). In addition, a separate instance of the T5-base model was specifically initialized as the value model for the PPO algorithm. Finally, we evaluate the trained LMs using three independent, highly accurate reward models, which assess various aspects of the LMs' outputs, including relevance, factuality, and completeness. A summary of the reward models' performance is provided in Table \ref{tab:reward_model_summary}.

\begin{table}[h!]
\footnotesize
\centering
\scalebox{1}{%
\begin{tabular}{@{}lcc@{}}
\toprule[1.0pt]
\textbf{Reward Model} & \textbf{Accuracy (\%)} & \textbf{F1 Score (\%)} \\ \midrule[1.0pt]
R$\phi_1$ (Relevance) & 69.6 & 68.5 \\ 
R$\phi_2$ (Factuality) & 77.8 & 67.5 \\ 
R$\phi_3$ (Completeness) & 70.9 & N/A \\ \bottomrule[1.0pt]
\end{tabular}
}
\caption{\small Summary of independent high-accuracy reward models used for evaluation.}
\label{tab:reward_model_summary}
\end{table}

A common pitfall in performing RLHF is reward gaming, where LMs maximize rewards in unintended ways, such as finding shortcuts in generation that attain high reward scores from the reward models, yet misalign with human preferences~\cite{pang2022reward}. To mitigate this, we following~\cite{wu2024fine} and set a KL threshold. When the divergence between the current policy and the reference policy exceeded this threshold, the training process was interrupted. This approach ensured that the model did not deviate excessively from the reference policy, effectively reducing the likelihood of reward manipulation.

\subsection{Are High-Accuracy and Deeply Trained Reward Models Always the Best?}

\paragraph{Setup.}

Building on the Basic Experimental Setup, reward models for relevance, factuality, and completeness from the QA-FEEDBACK dataset were used in PPO training. Performance was assessed at regular intervals, and top-performing instances were identified and visualized in three-dimensional plots.

\paragraph{Results.}

\begin{figure}[h!]
    \centering
    \includegraphics[width=0.95\linewidth]{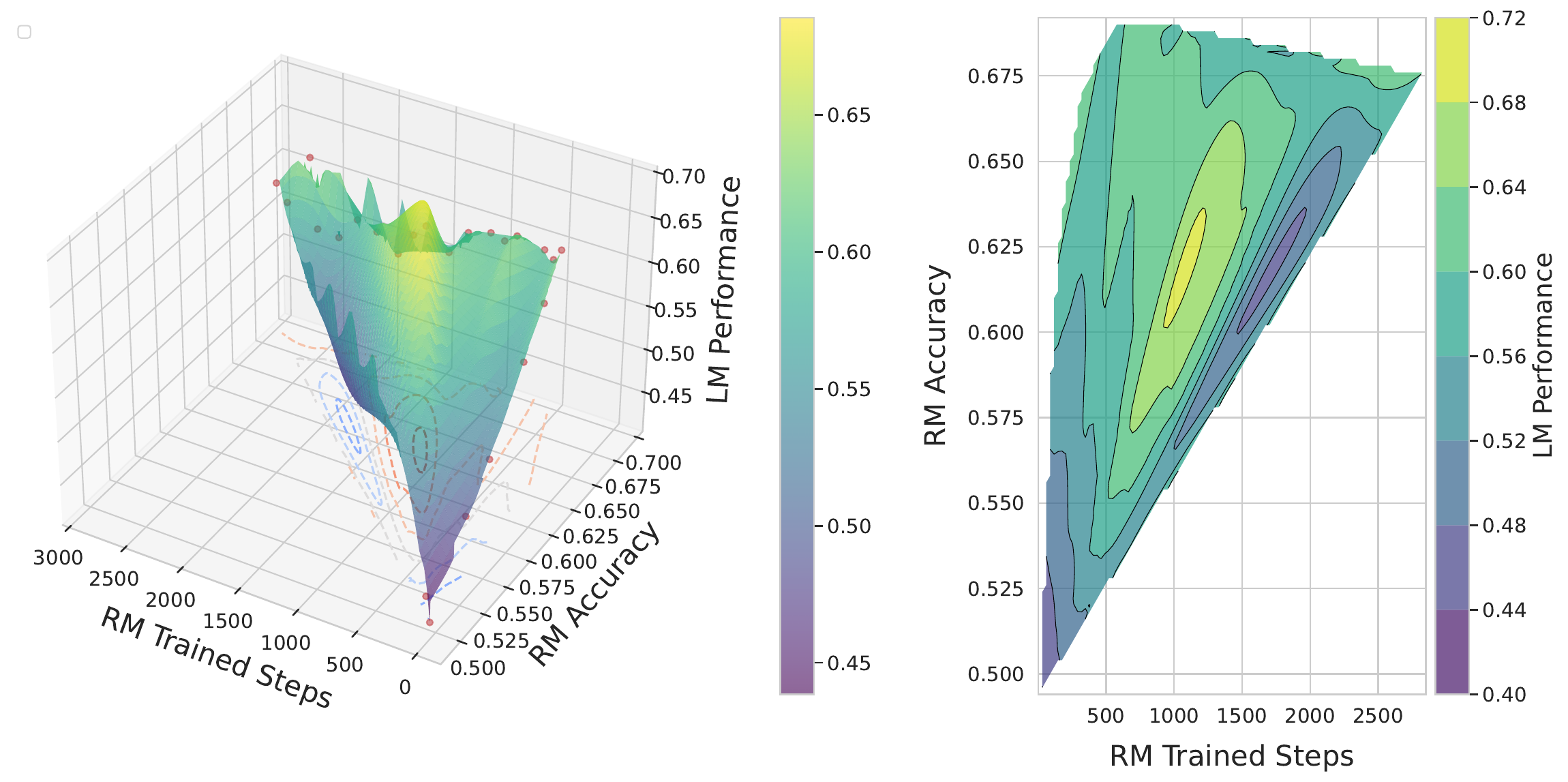}
    \caption{\small 3D surface plot evaluating relevance ratios for T5-small. Optimal performance was achieved with reward models having moderate accuracy.}
    \label{fig:3D-Surface-eval-rm-relevance-ratios-T5-small}
\end{figure}

\begin{figure}[h!]
    \centering
    \includegraphics[width=0.95\linewidth]{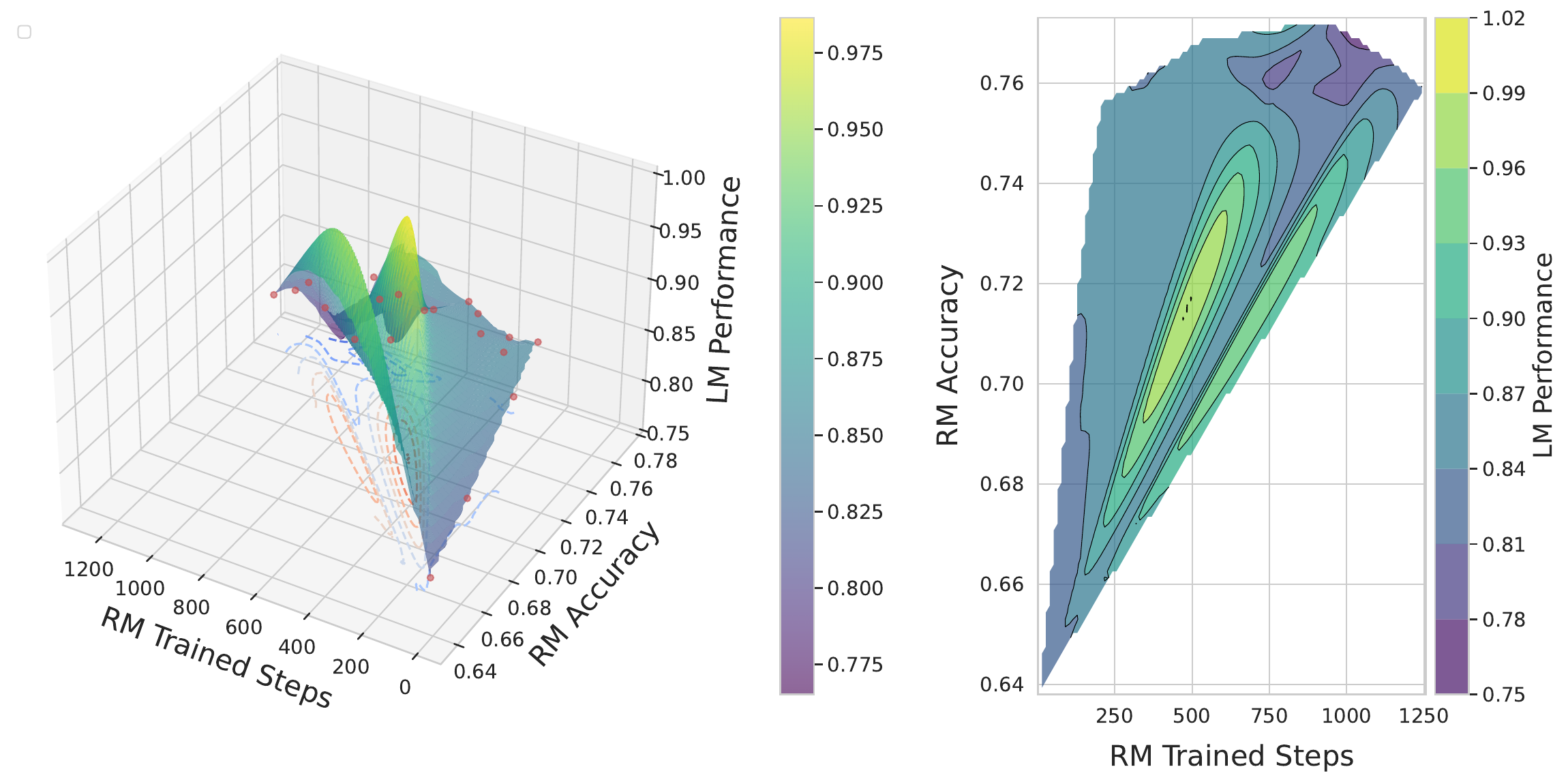}
    \caption{\small 3D surface plot evaluating factuality ratios for T5-small. The best performance was seen with reward models of moderate accuracy.}
    \label{fig:3D-Surface-eval-rm-factuality-ratios-T5-small}
\end{figure}

\begin{figure}[h!]
    \centering
    \includegraphics[width=0.95\linewidth]{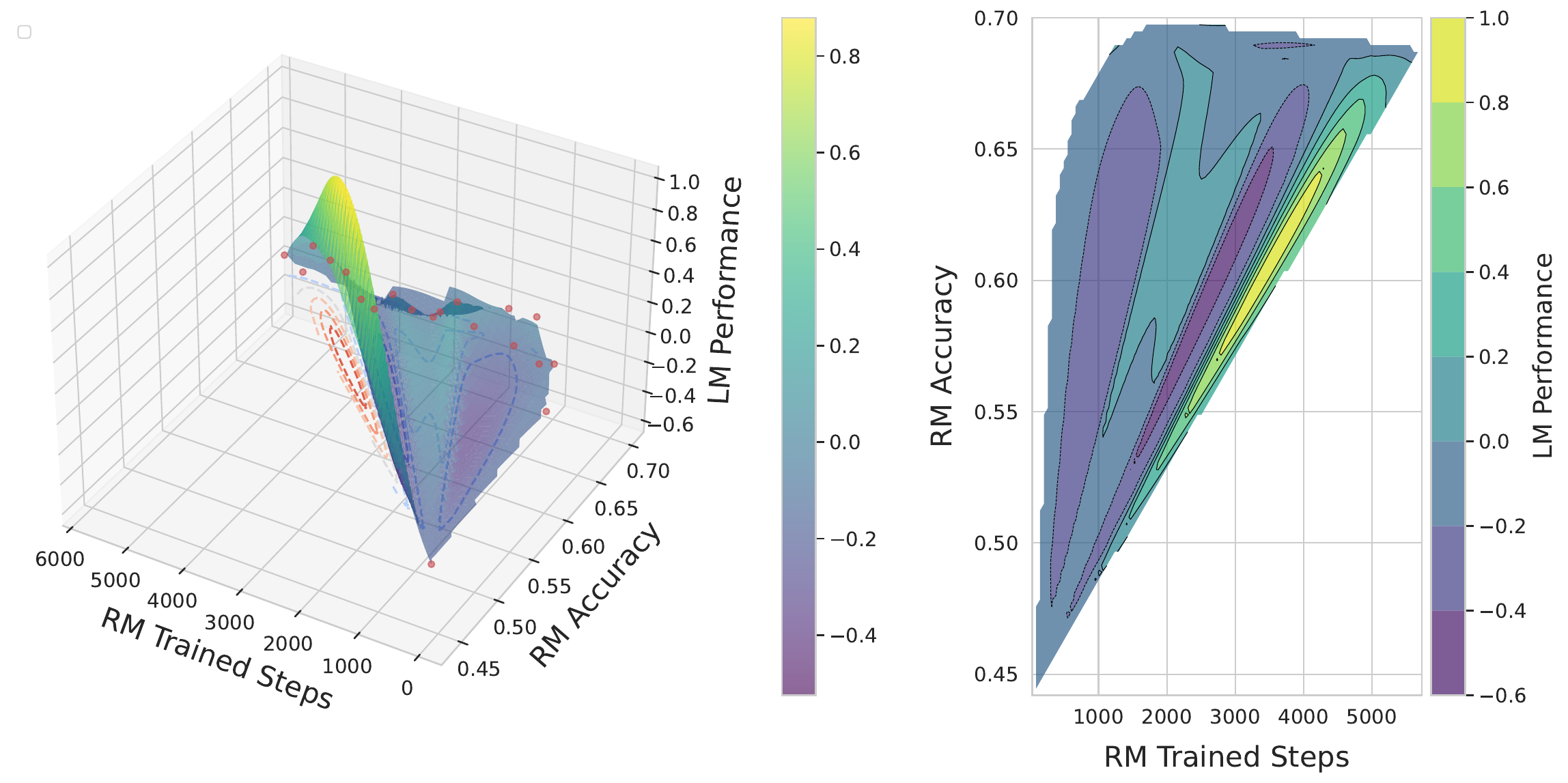}
    \caption{\small 3D surface plot evaluating completeness rewards for T5-small. Intermediate reward model strength yielded the best language model performance.}
    \label{fig:3D-Surface-eval-rm-completeness-rewards-T5-small}
\end{figure}

Figures \ref{fig:3D-Surface-eval-rm-relevance-ratios-T5-small} to \ref{fig:3D-Surface-eval-rm-completeness-rewards-T5-small} show that optimal language model performance is achieved using reward models with moderate accuracy and an appropriate number of trained steps. For the relevance task, the T5-small model performed best with moderately accurate reward models, effectively mitigating the risk of overfitting. Similarly, the results for factuality emphasized the importance of maintaining balanced reward model accuracy to prevent overfitting and ensure reliable outcomes. These findings suggest that overly accurate reward models can result in overfitting, which impairs the generalization ability of LMs. These trends were consistent across the T5-base and T5-large models, further supporting the conclusion that moderate accuracy in reward models strikes the best balance between training stability and performance. Detailed results for T5-base and T5-large are available in the Appendix \ref{appendix:3ds}.

\subsection{How Do Best and Most Accurate Reward Models Differ?}

\paragraph{Setup.}

This evaluation utilized three models: T5-small, T5-base, and T5-large, to compare the best-performing and most accurate reward models across relevance, factuality, and completeness tasks. The analysis focused on understanding the differences in reward behavior during training for each model. While the primary analysis in this section is based on the T5-small model, similar trends were observed with the T5-base and T5-large models, whose results are provided in the appendix for reference.

\paragraph{Results.}

\begin{figure}[h!]
    \centering
    \includegraphics[width=0.95\linewidth]{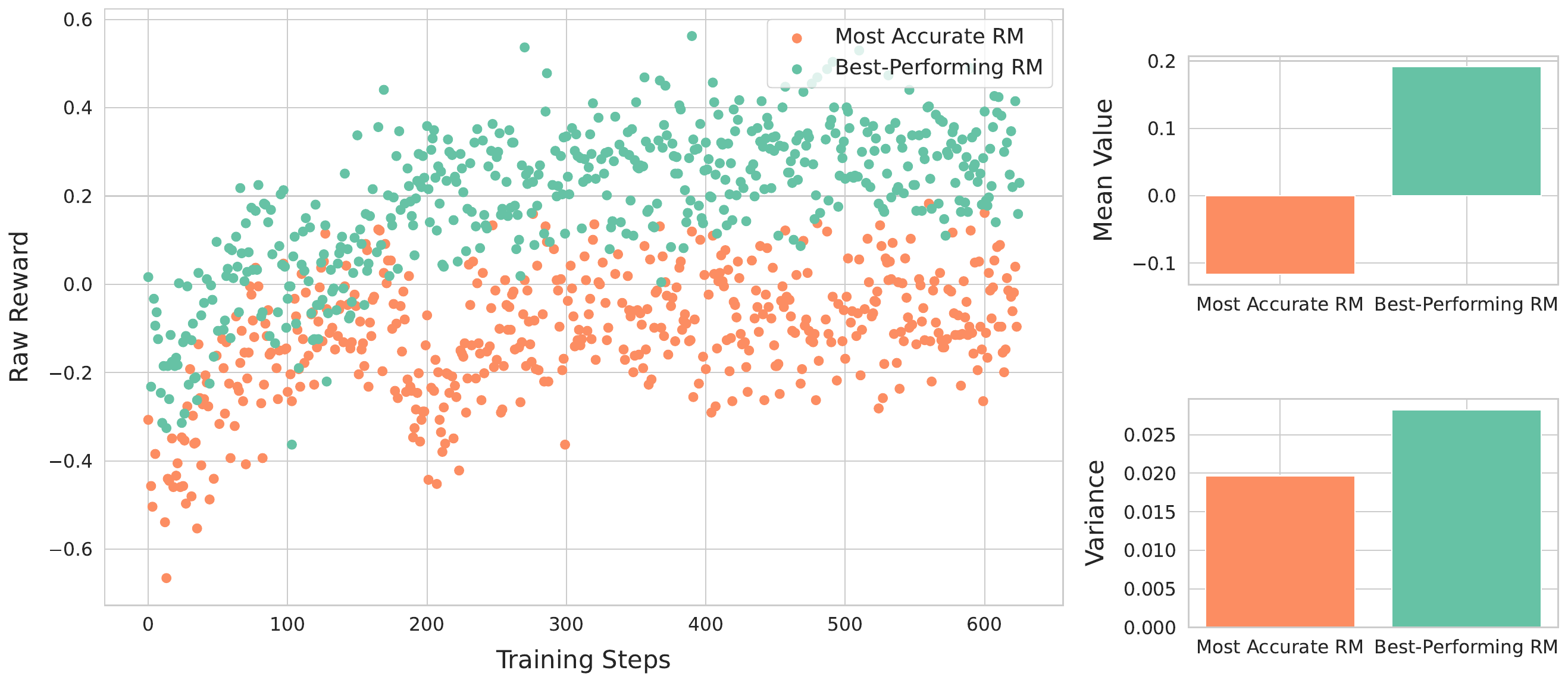}
    \caption{\small Reward analysis for relevance task (T5-small model): training steps vs. rewards (left), mean and variance of rewards (right).}
    \label{fig:rel-reward-analysis-T5-small}
\end{figure}

\begin{figure}[h!]
    \centering
    \includegraphics[width=0.95\linewidth]{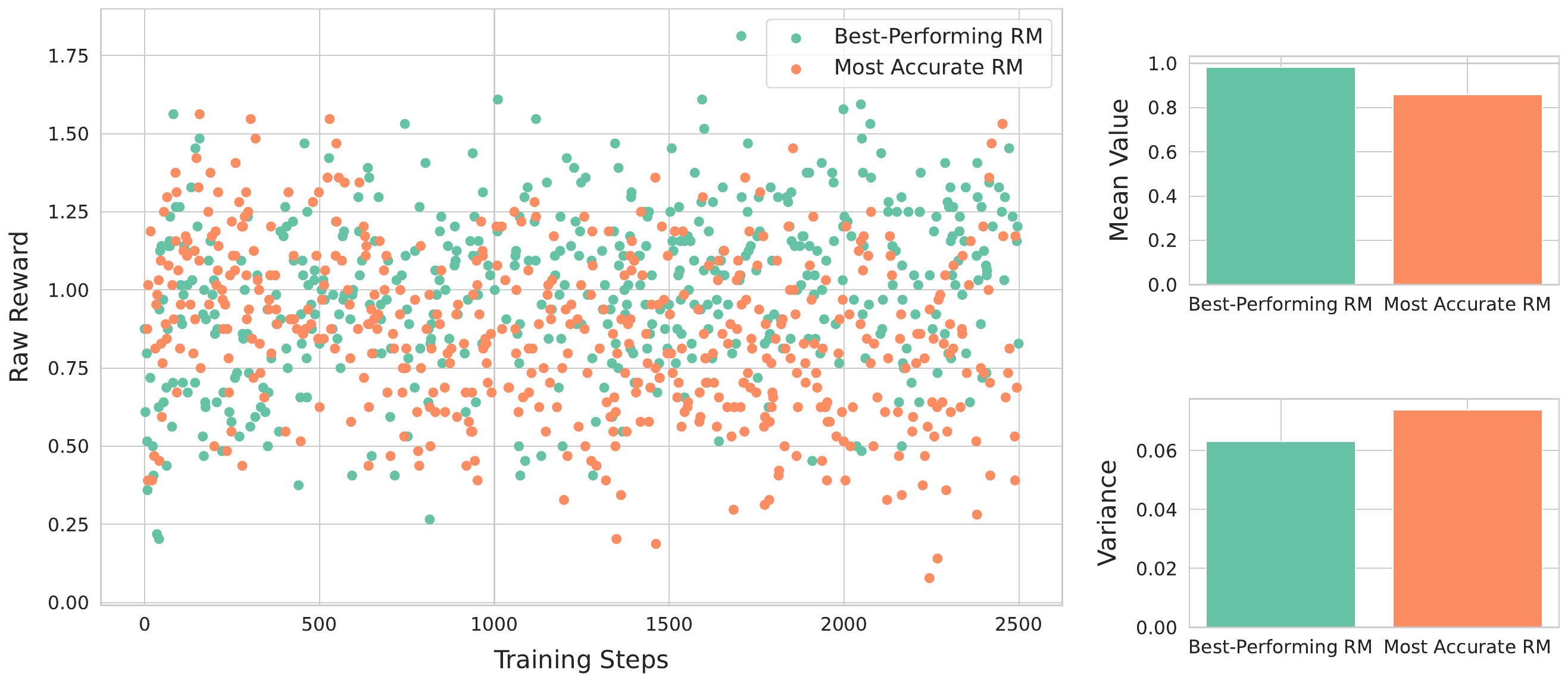}
    \caption{\small Reward analysis for factuality task (T5-small model): training steps vs. rewards (left), mean and variance of rewards (right).}
    \label{fig:fact-reward-analysis-T5-small}
\end{figure}

\begin{figure}[h!]
    \centering
    \includegraphics[width=0.95\linewidth]{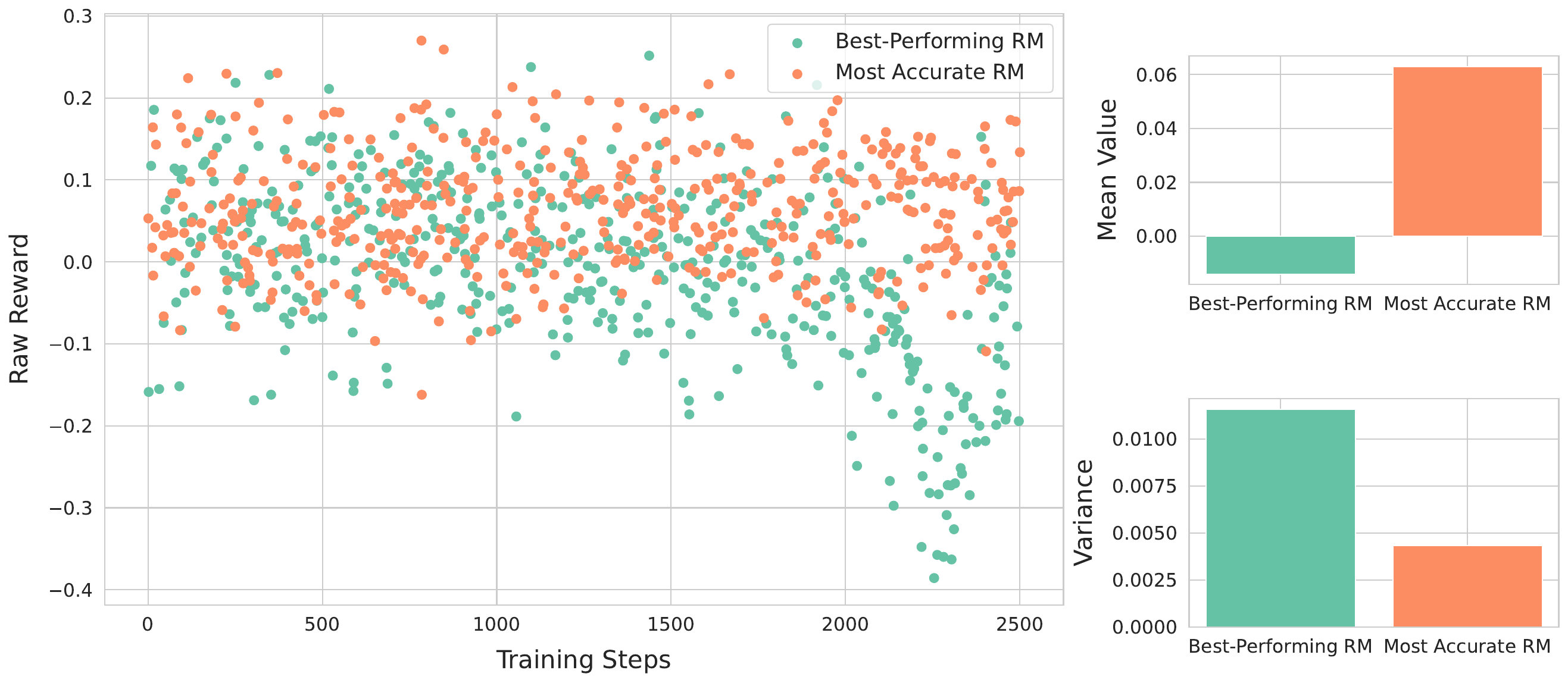}
    \caption{\small Reward analysis for completeness task (T5-small model): training steps vs. rewards (left), mean and variance of rewards (right).}
    \label{fig:comp-reward-analysis-T5-small}
\end{figure}

Figures \ref{fig:rel-reward-analysis-T5-small}, \ref{fig:fact-reward-analysis-T5-small}, and \ref{fig:comp-reward-analysis-T5-small} illustrate the distinct strategies of the best-performing reward models compared to the most accurate models using the T5-small model. For the relevance task, the best-performing reward model provided higher and more variable rewards (Figure \ref{fig:rel-reward-analysis-T5-small}), indicating an aggressive approach that likely stimulated the generation of more relevant outputs. In the factuality task, this model maintained higher mean rewards with less variability (Figure \ref{fig:fact-reward-analysis-T5-small}), promoting factual accuracy. Conversely, for the completeness task, it employed a conservative strategy with lower average rewards but greater variability (Figure \ref{fig:comp-reward-analysis-T5-small}).

\paragraph{Analysis.}

Moderately accurate best-performing reward models typically align rewards with task requirements. In both relevance and factuality tasks, these models provide higher and more varied rewards, thus encouraging the generation of more relevant and accurate outputs. This variability allows LMs to explore a broader range of responses, improving the quality of the generated text. Conversely, in completeness tasks, a conservative strategy with lower average rewards but greater variability helps ensure thorough and comprehensive text evaluation. The trends observed in T5-small models are consistent with those seen in T5-base and T5-large models, further supporting the conclusion that moderate accuracy in reward models effectively balances overfitting and underfitting. Detailed results for T5-base and T5-large can be found in the Appendix \ref{appendix:rewardA}.

\subsection{How Do Best and Most Accurate Rewards Impact Models?}

\paragraph{Setup.}

This section evaluates the impact of reward models on the training dynamics of T5-small, T5-base, and T5-large models in relevance, factuality, and completeness tasks, with a focus on KL divergence trends to assess stability and adaptability. While the results presented here focus on the T5-small model, similar trends were observed for the T5-base and T5-large models, whose results are provided in the appendix.

\paragraph{KL Divergence and Its Role in RLHF}
KL divergence (Kullback-Leibler divergence) is a measure of how one probability distribution \(P\) diverges from a second, expected probability distribution \(Q\)~\cite{kullback1951information}. It is commonly used in reinforcement learning to constrain the difference between the current policy and a reference policy during training. Mathematically, KL divergence is defined as:

\begin{equation}
D_{\mathrm{KL}}(P \parallel Q) = \sum_{i} P(i) \log \left( \frac{P(i)}{Q(i)} \right)
\end{equation}

In the context of RLHF, KL divergence serves as a regularization term to prevent the trained policy from deviating excessively from the reference policy. This constraint helps to stabilize the training process by reducing the chance of reward hacking or reward gaming, where the model could exploit the reward system without truly improving performance~\cite{pang2022reward}.

\paragraph{Results.}

\begin{figure}[h!]
    \centering
    \includegraphics[width=0.95\linewidth]{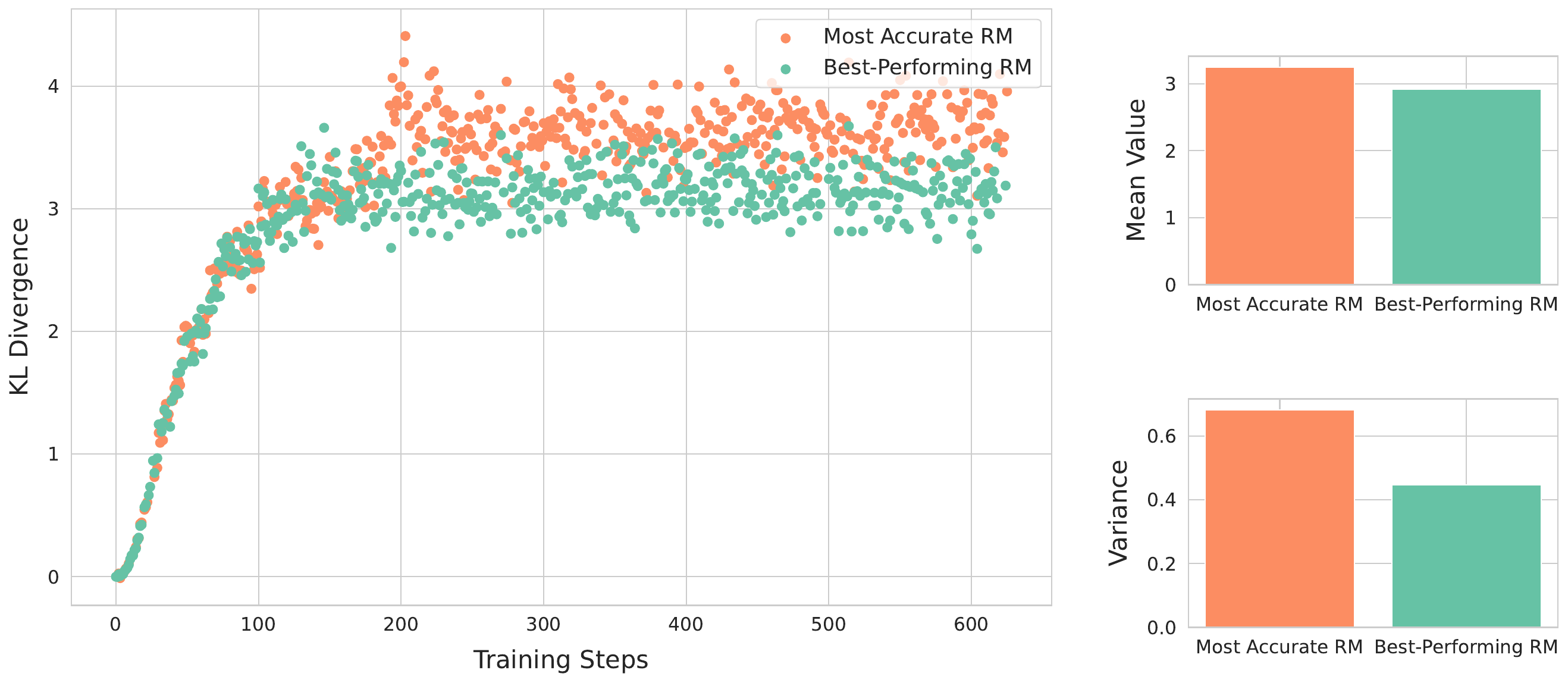}
    \caption{\small Relevance task KL divergence (T5-small model): training steps vs. KL divergence (left), mean and variance of rewards (right).}
    \label{fig:rel-reward-analysis-3-T5-small}
\end{figure}

\begin{figure}[h!]
    \centering
    \includegraphics[width=0.95\linewidth]{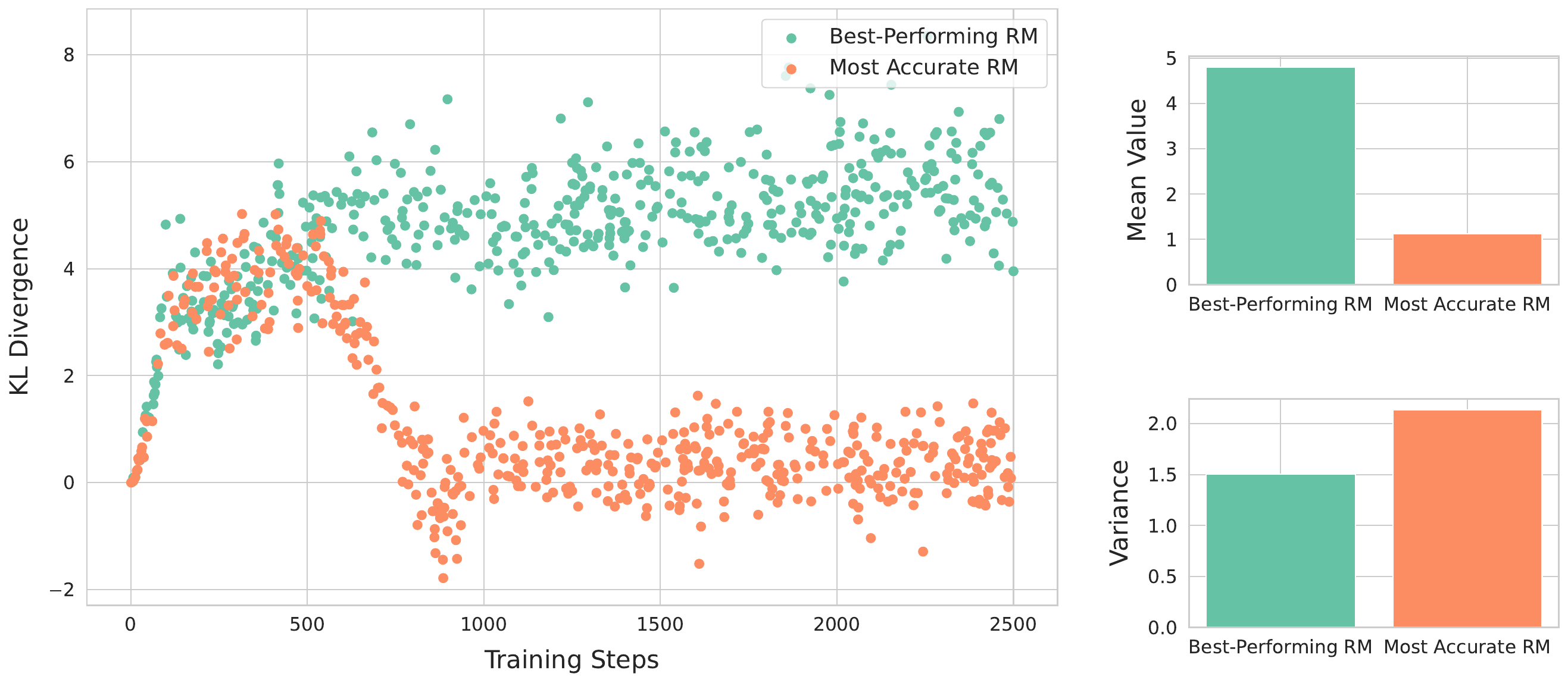}
    \caption{\small Factuality task KL divergence (T5-small model): training steps vs. KL divergence (left), mean and variance of rewards (right).}
    \label{fig:fact-reward-analysis-3-T5-small}
\end{figure}

\begin{figure}[h!]
    \centering
    \includegraphics[width=0.95\linewidth]{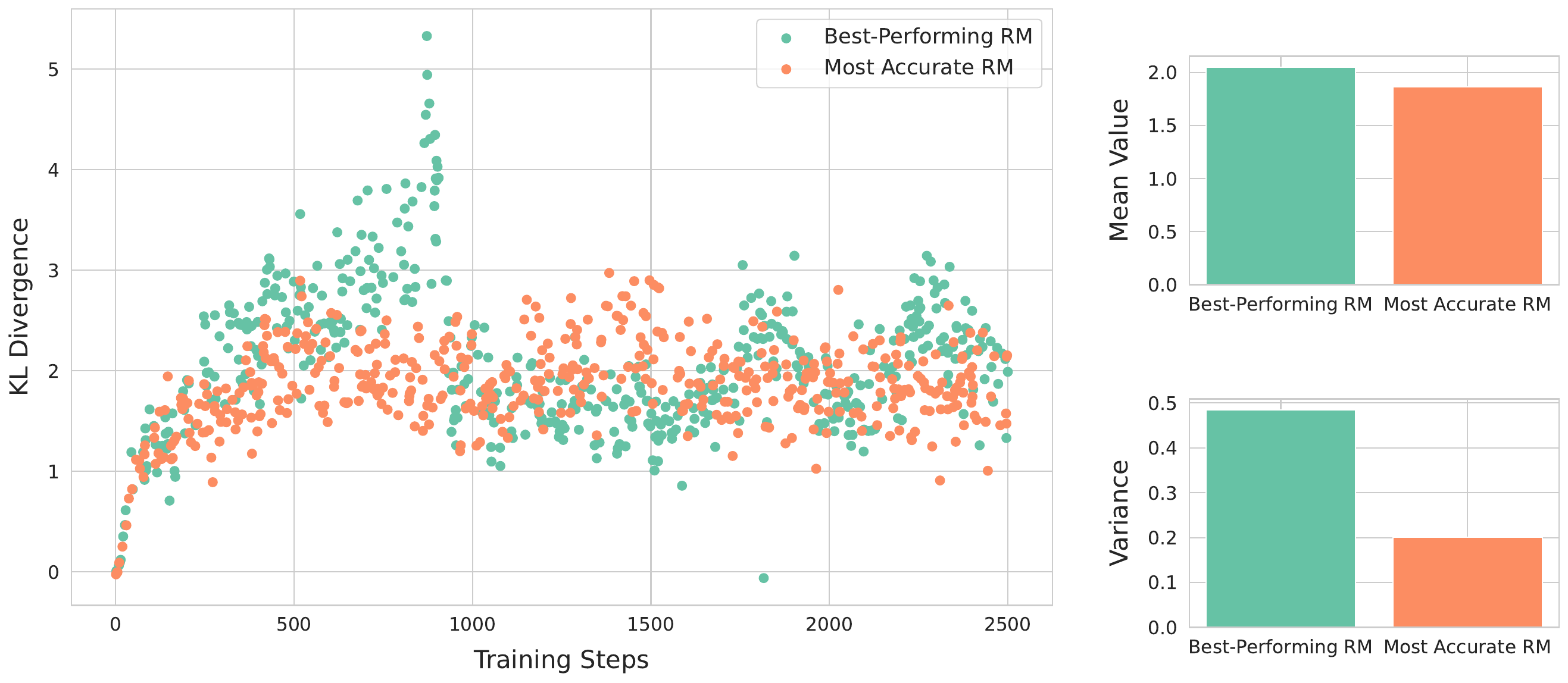}
    \caption{\small Completeness task KL divergence (T5-small model): training steps vs. KL divergence (left), mean and variance of rewards (right).}
    \label{fig:comp-reward-analysis-3-T5-small}
\end{figure}

Comparing KL divergence trends revealed significant differences in how LMs aligned with the training data. For the relevance task, the best reward model resulted in consistently lower KL divergence and variance, indicating stable alignment (Figure \ref{fig:rel-reward-analysis-3-T5-small}). In the factuality task, the best reward model exhibited higher mean KL divergence but lower variance, suggesting a consistent yet varied alignment process (Figure \ref{fig:fact-reward-analysis-3-T5-small}). For the completeness task, the best reward model showed higher mean and variance in KL divergence, indicating a flexible approach suitable for evaluating complex texts (Figure \ref{fig:comp-reward-analysis-3-T5-small}).

\paragraph{Analysis.}

Best-performing reward models, which are typically of moderate accuracy, create a balanced training environment that facilitates both stability and adaptability. In relevance and factuality tasks, these models encourage stable learning, enhancing the relevance and accuracy of outputs. For the completeness task, the flexibility in handling complex texts is demonstrated by higher variance in KL divergence. The observed trends in T5-small models were consistent with those seen in T5-base and T5-large models, further validating the conclusion that moderate accuracy in reward models effectively balances overfitting and underfitting. Detailed results for T5-base and T5-large models can be found in the Appendix \ref{appendix:klD}.

% 总结
\section{Conclusion and Future Work}

This study demonstrates that LMs trained with moderately accurate reward models in RLHF achieve optimal performance, challenging the conventional belief that higher accuracy is always more beneficial. The results show that moderately accurate reward models offer more task-aligned feedback and foster a balanced, stable training process, promoting better generalization. This research highlights the limitations of relying exclusively on highly accurate reward models, as excessive focus on accuracy may lead to suboptimal outcomes. In future work, it will be crucial to further explore the potential overfitting of reward models, particularly in their ability to generalize to out-of-distribution (OOD) tasks. Techniques such as regularization, data augmentation, and explicit OOD evaluation will be key areas of investigation to enhance the robustness of reward models across diverse scenarios and ensure their effectiveness in guiding LMs in broader, more complex NLP tasks.

% 局限性
\section*{Limitations}

\paragraph{Dataset Constraints.}
The conclusions are drawn from the QA-FEEDBACK dataset~\cite{wu2024fine}, which is specialized in generating long-form responses to factual inquiries. This focus may limit the generalizability of the results, necessitating validation across various datasets, including those pertaining to conversational and question-answering contexts.

\paragraph{Model Scope.}
The evaluation utilized T5 models of different scales for initial validation~\cite{raffel2020exploring}. Future investigations should incorporate more complex models, such as Llama2~\cite{touvron2023llama}, to gain deeper insights and verify the robustness of the proposed methodologies across a broader range of model architectures.

\paragraph{Reward Model Variations.}
This study did not explore the impact of different reward model sizes and architectures on RLHF performance. The reward models used were based on a single architecture, which may limit the applicability of the findings. Future research should systematically investigate how variations in reward model size, capacity, and design affect the learning process, generalization, and overall RLHF performance, particularly in diverse NLP tasks. Understanding the influence of these factors will be crucial for developing more robust and scalable reward models that can generalize across a wider range of applications.

\section*{Acknowledgements}
We thank Xingluan (AI Cloud computing service), EIT and IDT High Performance Computing Center for providing computational resources for this project. This work is supported by 2035 Key Research and Development Program of Ningbo City under Grant No.2024Z127.

\bibliography{custom}

\appendix

\section{Supplementary 3D Surface Plots}
\FloatBarrier
\label{appendix:3ds}
\begin{figure}[H]
    \centering
    \includegraphics[width=0.95\linewidth]{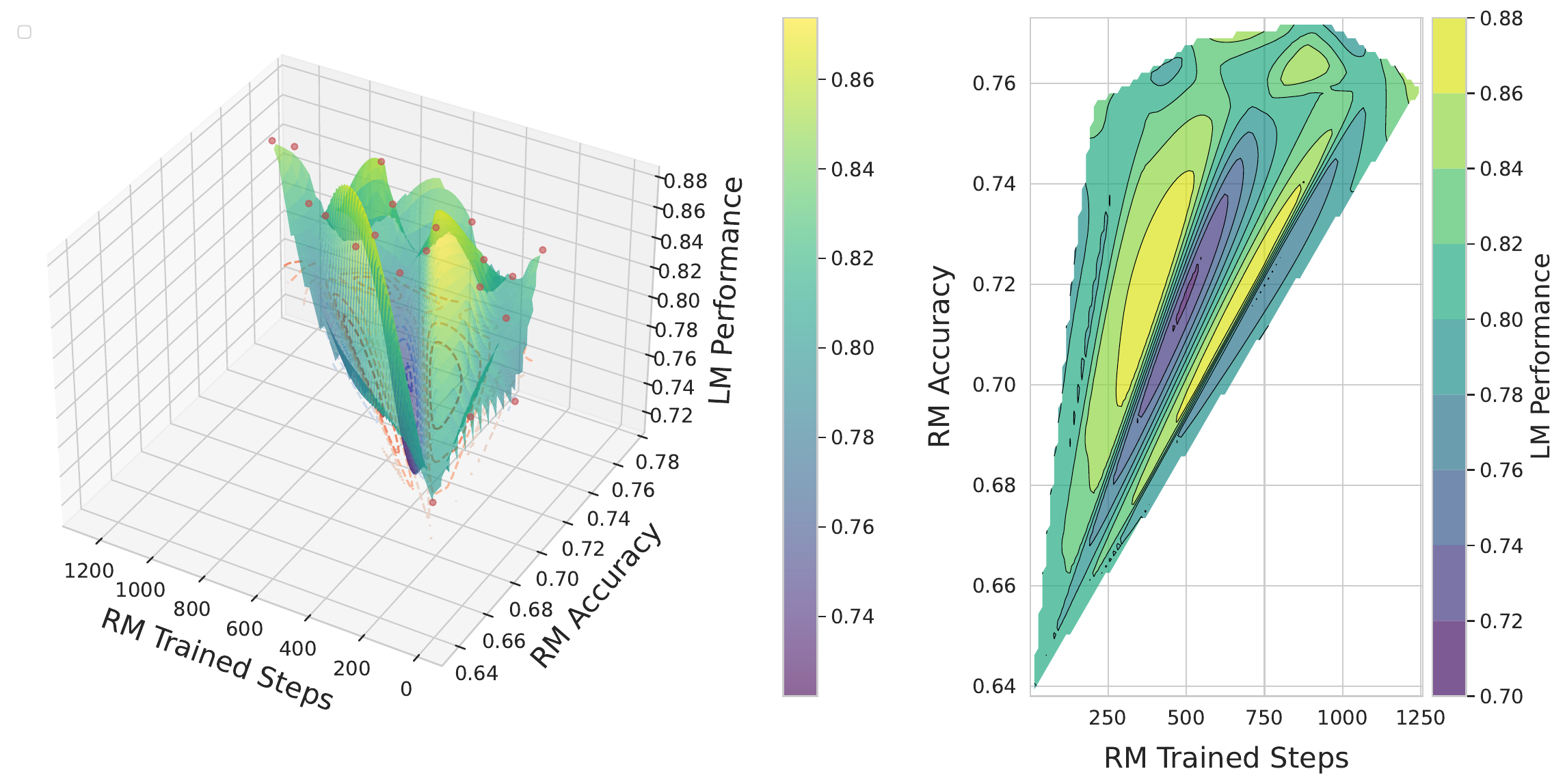}
    \caption{\small 3D surface plot evaluating factuality ratios for T5-base. Optimal performance was achieved with reward models having moderate accuracy.}
    % \label{fig:3D-Surface-eval-rm-factuality-ratios-T5-base}
\end{figure}

\begin{figure}[H]
    \centering
    \includegraphics[width=0.95\linewidth]{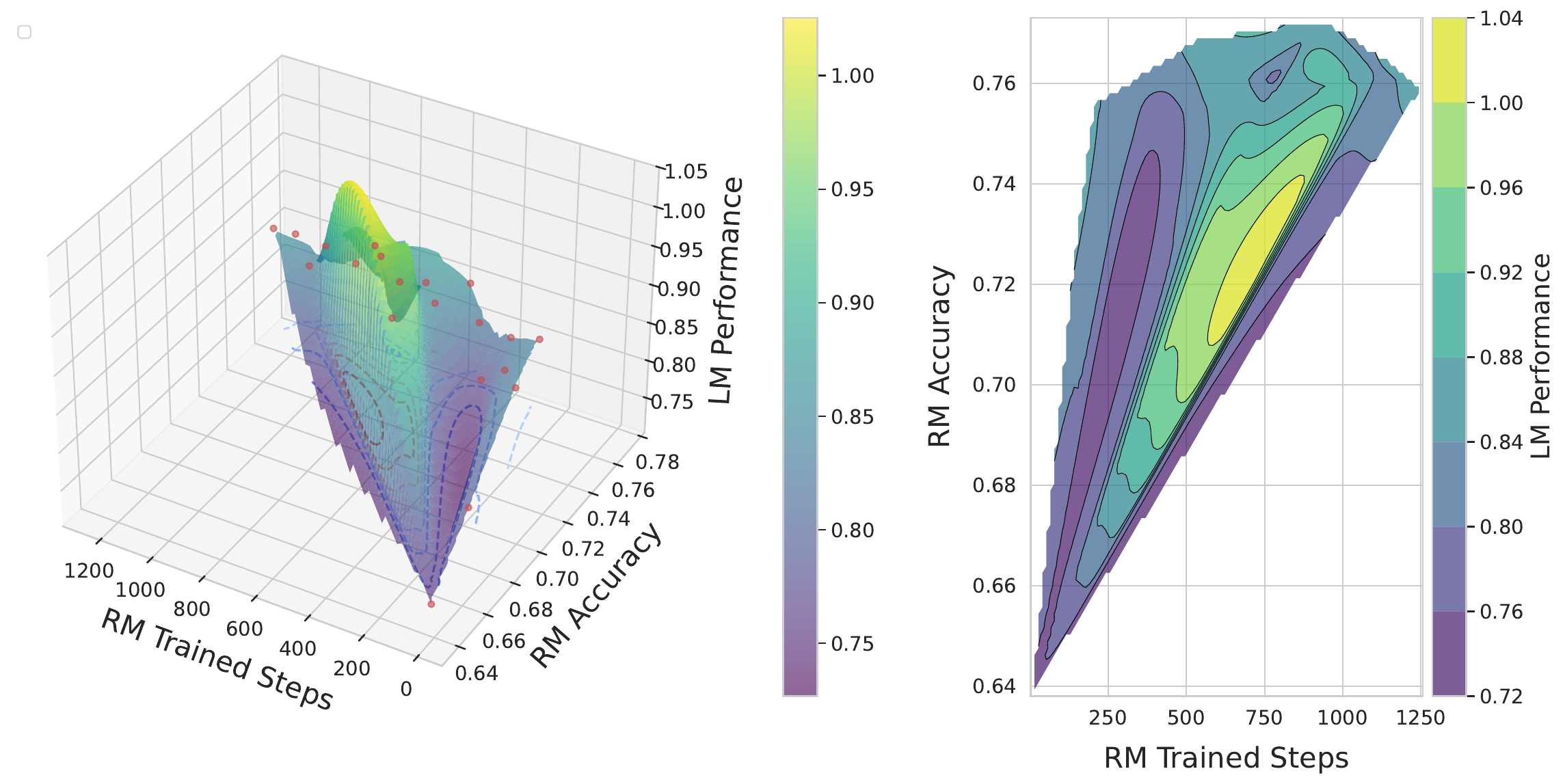}
    \caption{\small 3D surface plot evaluating factuality ratios for T5-large. The best performance was seen with reward models of moderate accuracy.}
    % \label{fig:3D-Surface-eval-rm-factuality-ratios-T5-large}
\end{figure}

\begin{figure}[H]
    \centering
    \includegraphics[width=0.95\linewidth]{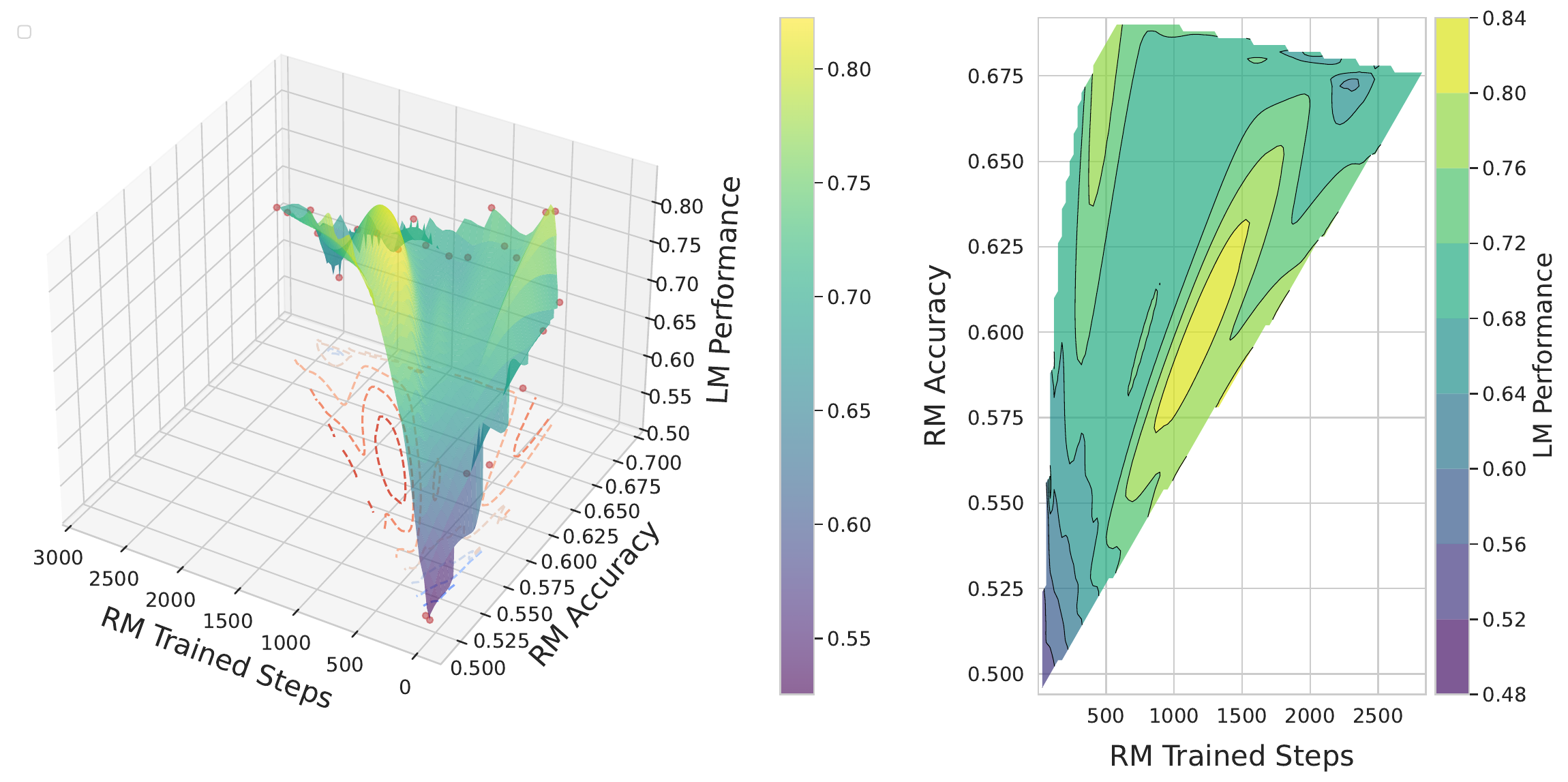}
    \caption{\small 3D surface plot evaluating relevance ratios for T5-base. Optimal performance was achieved with reward models having moderate accuracy.}
    % \label{fig:3D-Surface-eval-rm-relevance-ratios-T5-base}
\end{figure}

\begin{figure}[H]
    \centering
    \includegraphics[width=0.95\linewidth]{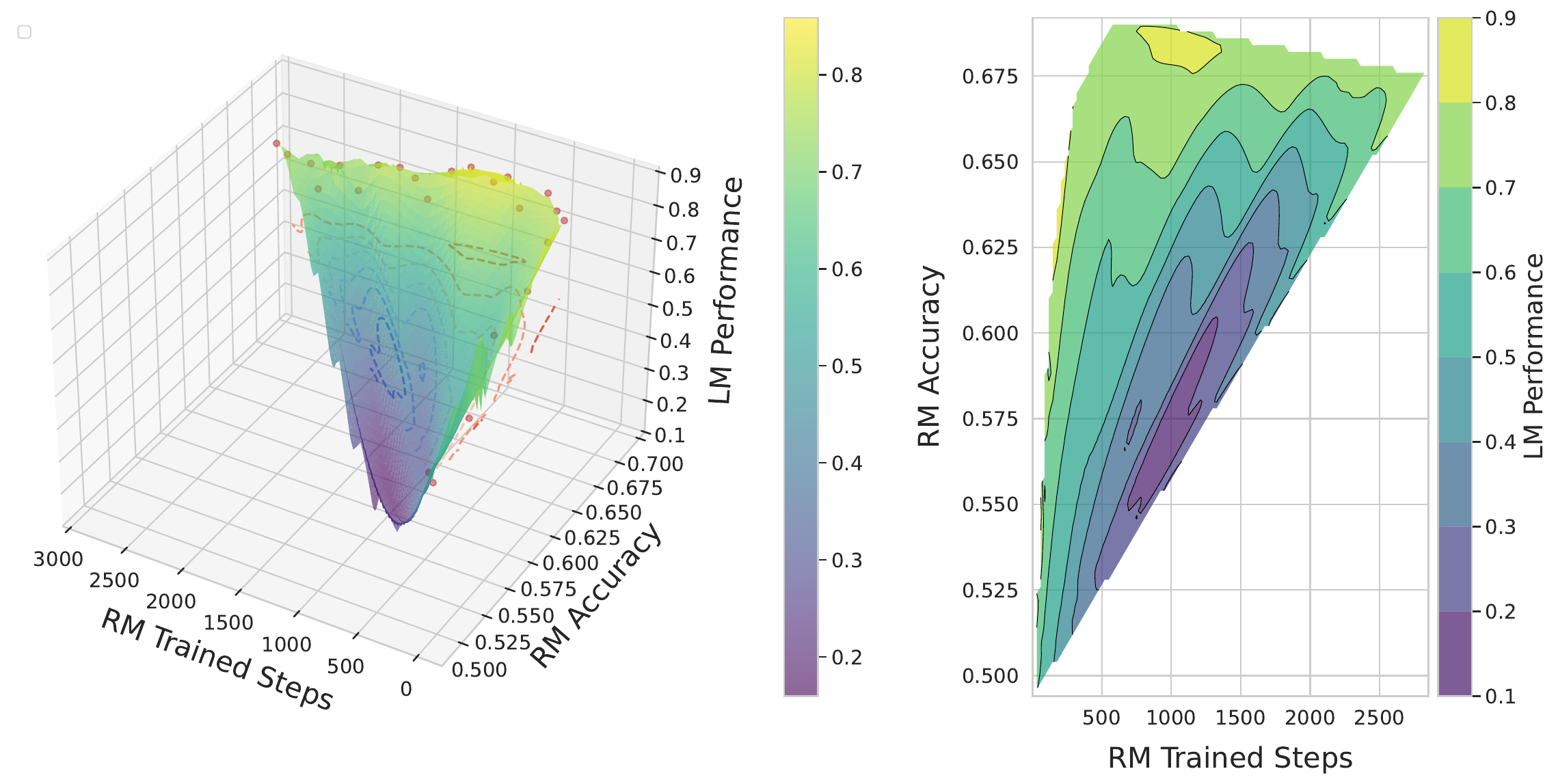}
    \caption{\small 3D surface plot evaluating relevance ratios for T5-large. Optimal performance was achieved with reward models having moderate accuracy.}
    % \label{fig:3D-Surface-eval-rm-relevance-ratios-T5-large}
\end{figure}

\begin{figure}[H]
    \centering
    \includegraphics[width=0.95\linewidth]{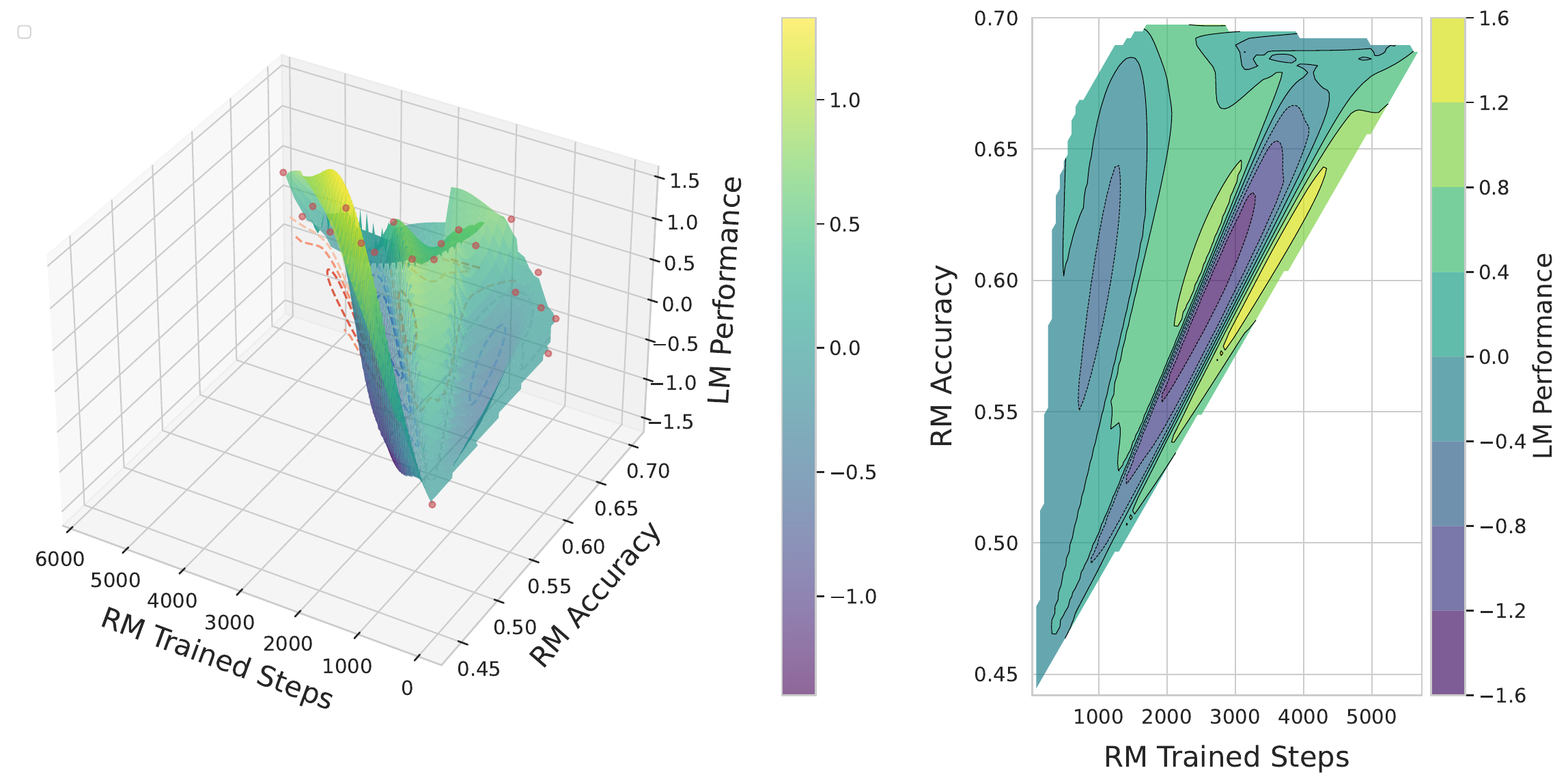}
    \caption{\small 3D surface plot evaluating completeness rewards for T5-base. Intermediate reward model strength yielded the best language model performance.}
    % \label{fig:3D-Surface-eval-rm-completeness-rewards-T5-base}
\end{figure}

\begin{figure}[H]
    \centering
    \includegraphics[width=0.95\linewidth]{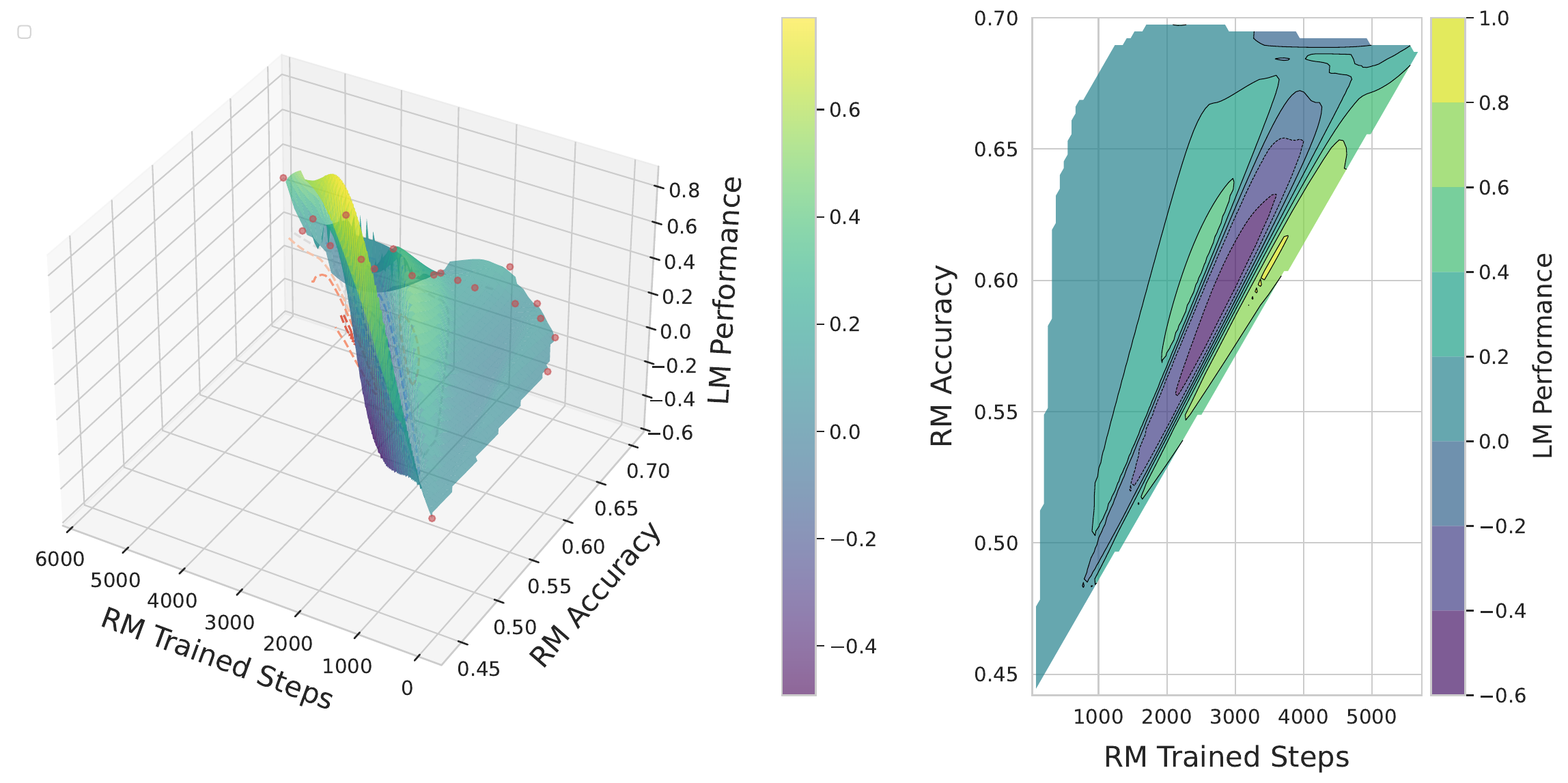}
    \caption{\small 3D surface plot evaluating completeness rewards for T5-large. Intermediate reward model strength yielded the best language model performance.}
    % \label{fig:3D-Surface-eval-rm-completeness-rewards-T5-large}
\end{figure}

\section{Supplementary Reward Analysis}
\FloatBarrier
\label{appendix:rewardA}
\begin{figure}[H]
    \centering
    \includegraphics[width=0.95\linewidth]{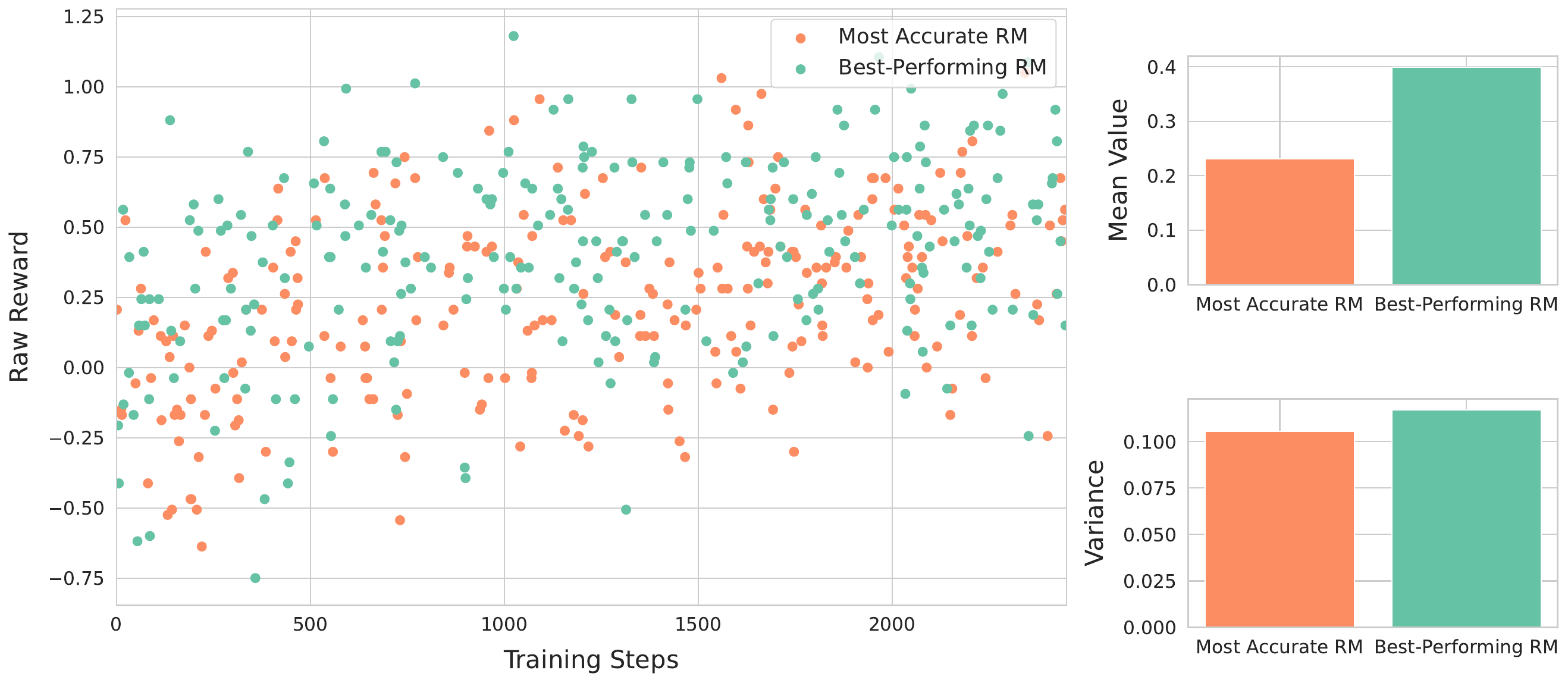}
    \caption{\small Reward analysis for relevance task (T5-base model): training steps vs. rewards (left), mean and variance of rewards (right).}
    % \label{fig:rel-reward-analysis}
\end{figure}

\begin{figure}[H]
    \centering
    \includegraphics[width=0.95\linewidth]{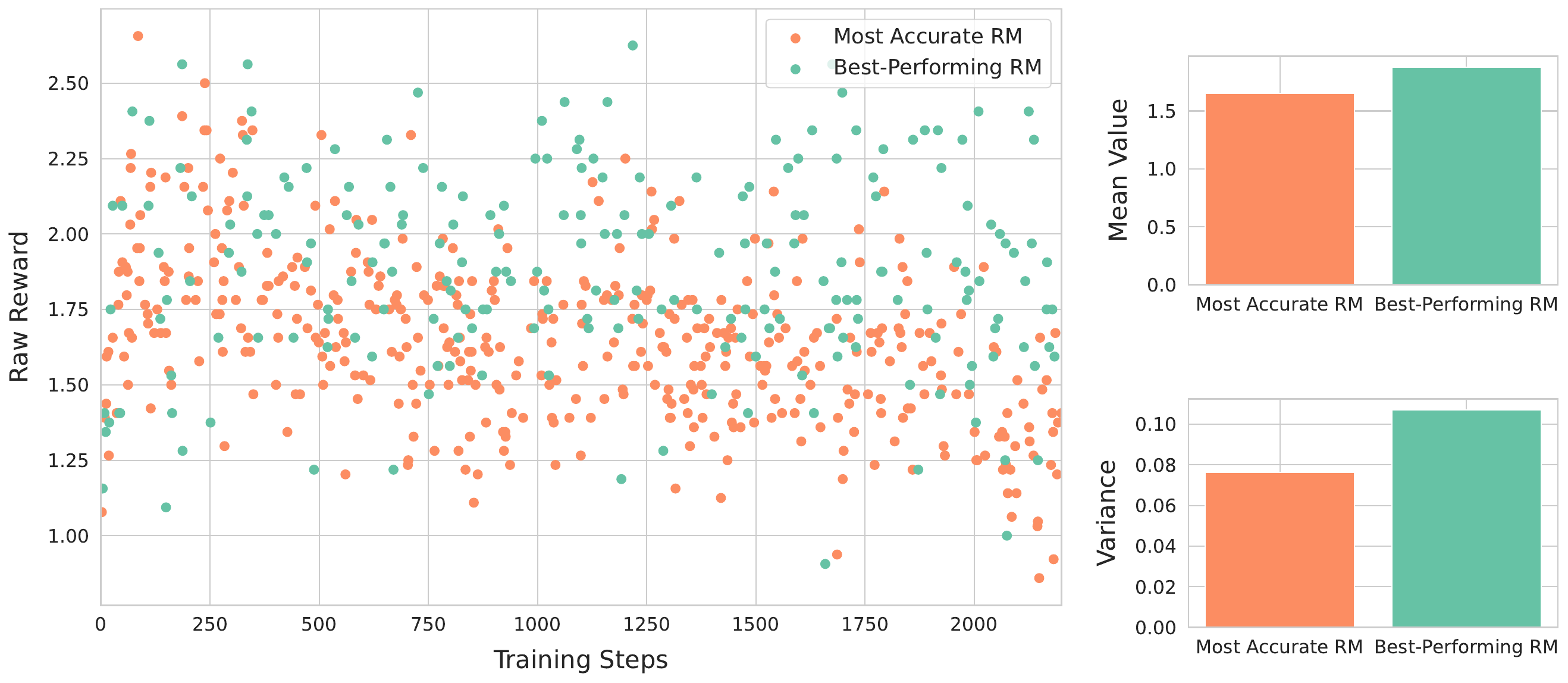}
    \caption{\small Reward analysis for factuality task (T5-base model): training steps vs. rewards (left), mean and variance of rewards (right).}
    % \label{fig:fact-reward-analysis}
\end{figure}

\begin{figure}[H]
    \centering
    \includegraphics[width=0.95\linewidth]{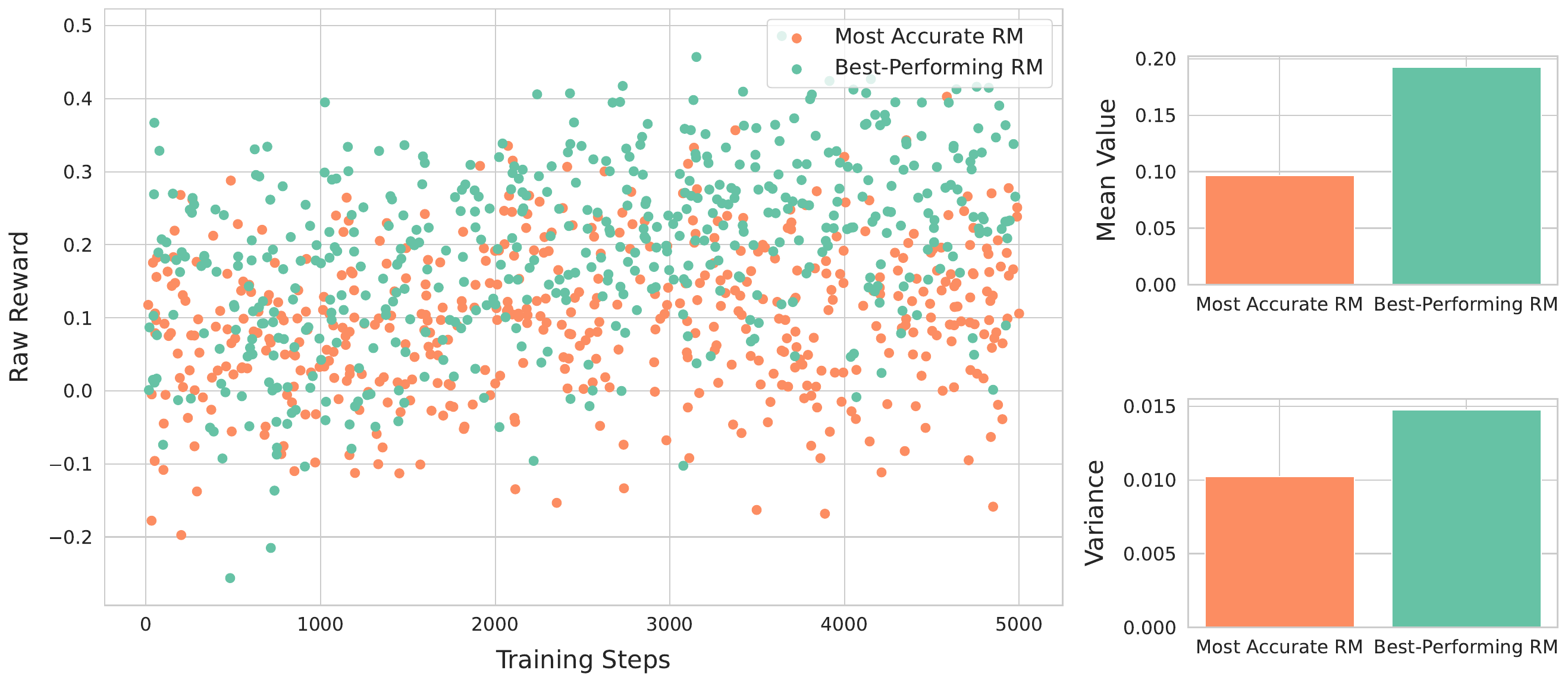}
    \caption{\small Reward analysis for completeness task (T5-base model): training steps vs. rewards (left), mean and variance of rewards (right).}
    % \label{fig:comp-reward-analysis}
\end{figure}

\begin{figure}[H]
    \centering
    \includegraphics[width=0.95\linewidth]{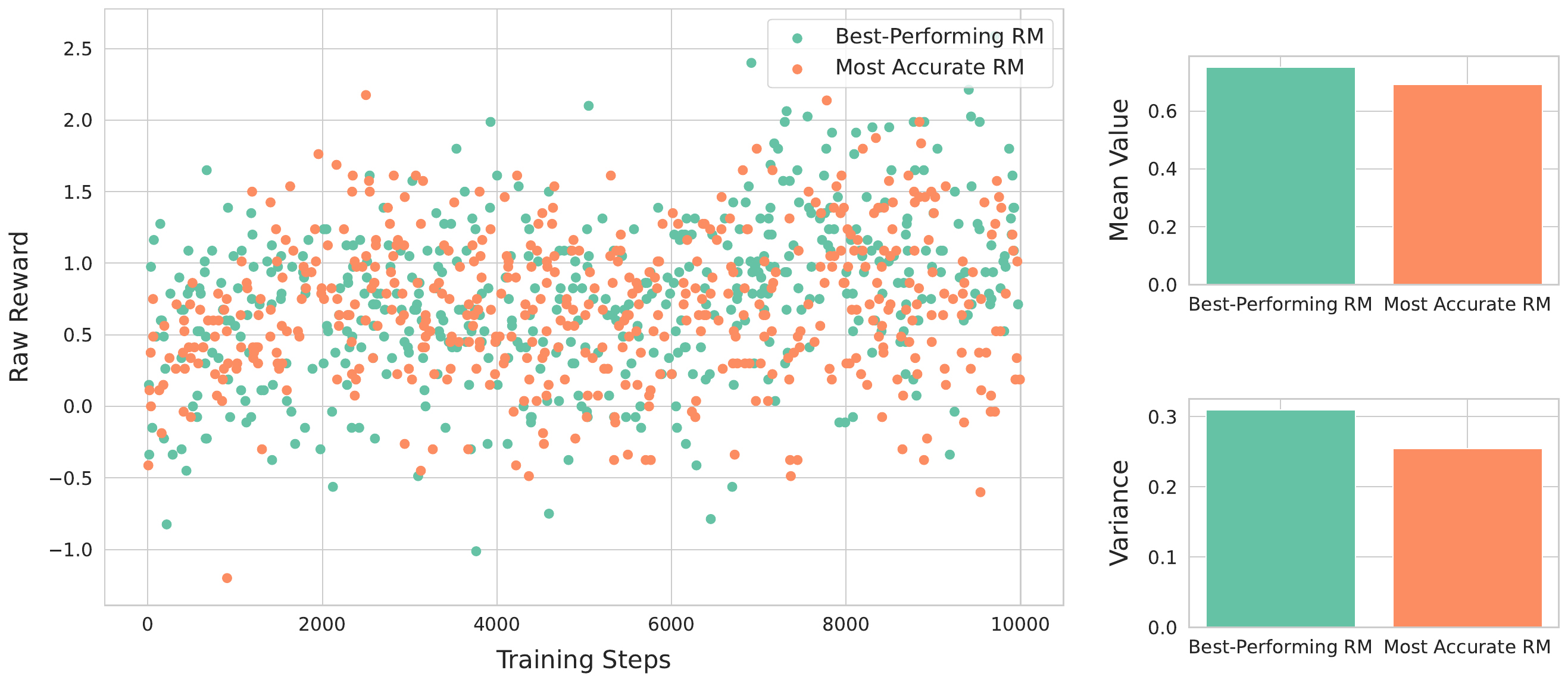}
    \caption{\small Reward analysis for relevance task (T5-large model): training steps vs. rewards (left), mean and variance of rewards (right).}
    % \label{fig:rel-reward-analysis}
\end{figure}

\begin{figure}[H]
    \centering
    \includegraphics[width=0.95\linewidth]{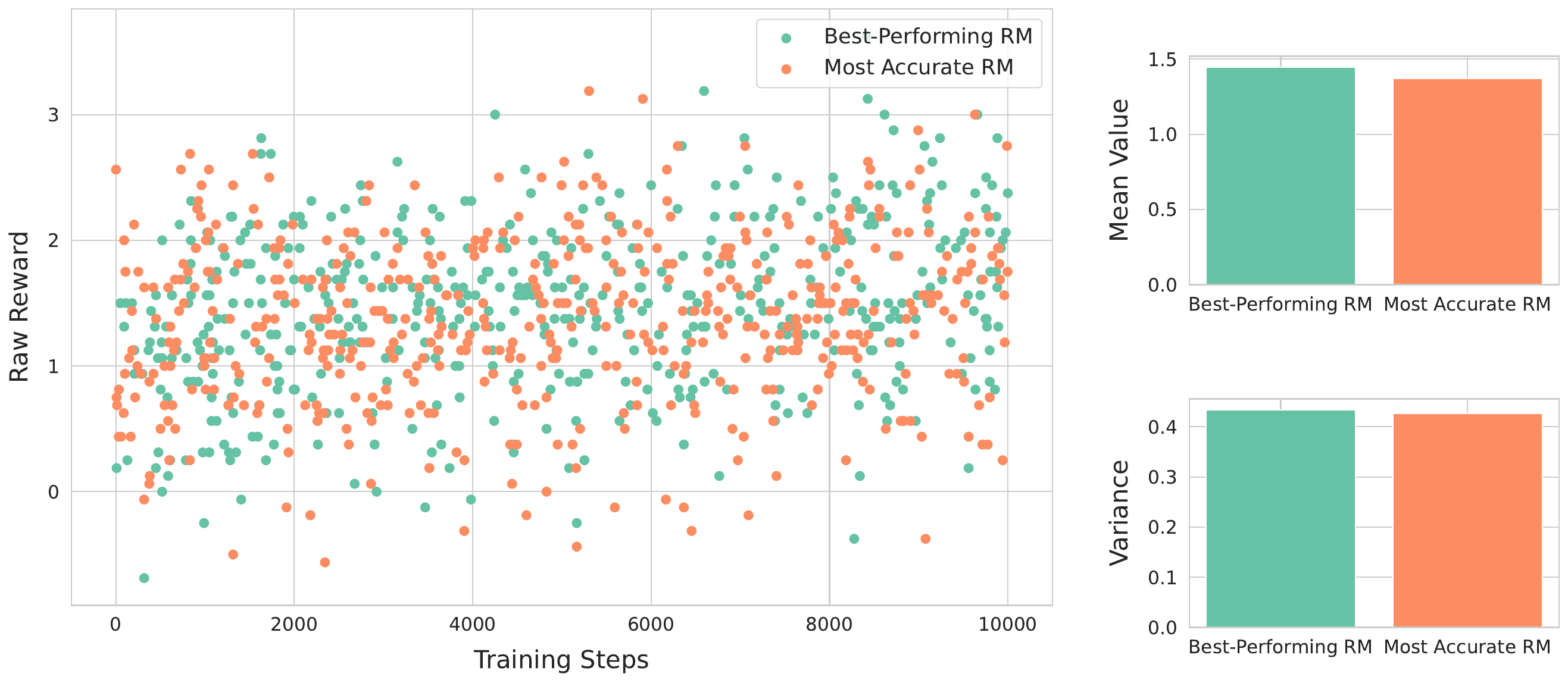}
    \caption{\small Reward analysis for factuality task (T5-large model): training steps vs. rewards (left), mean and variance of rewards (right).}
    % \label{fig:fact-reward-analysis}
\end{figure}

\begin{figure}[H]
    \centering
    \includegraphics[width=0.95\linewidth]{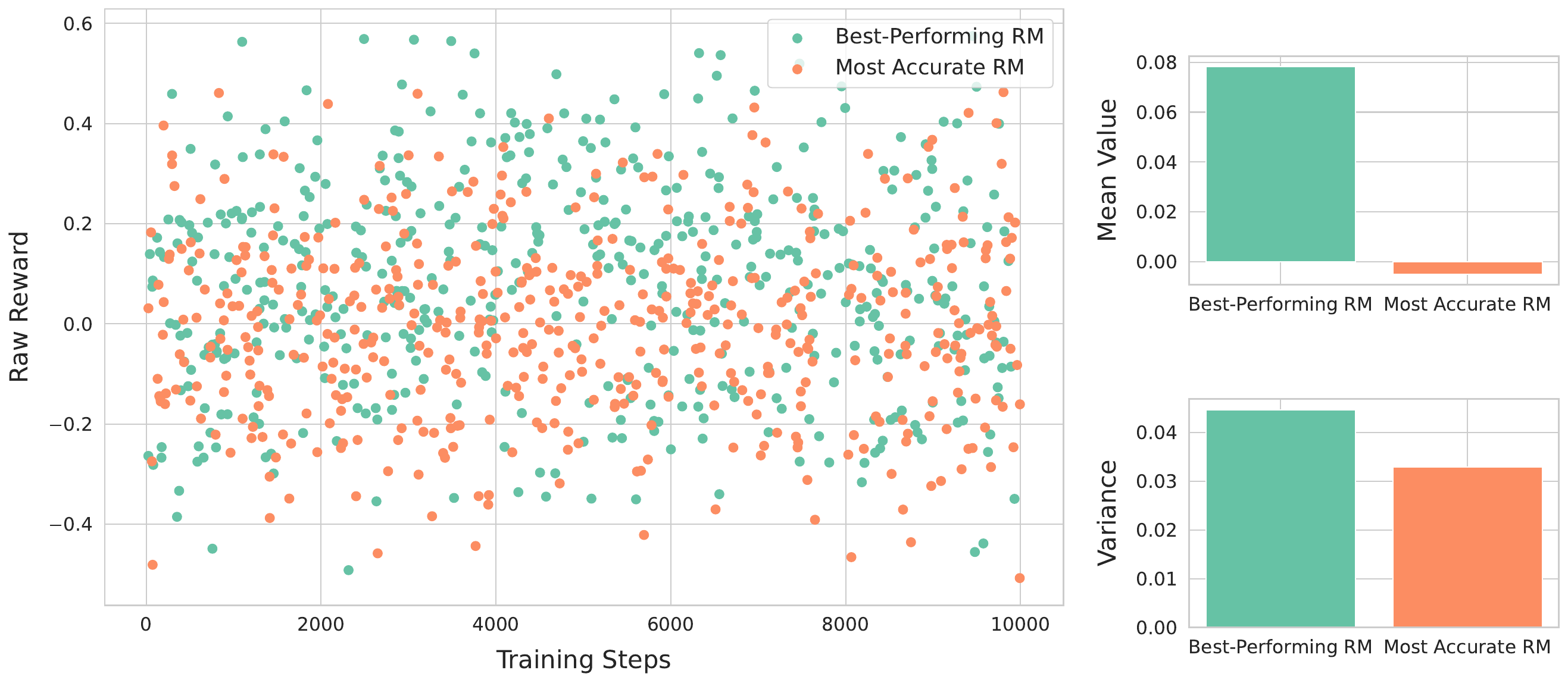}
    \caption{\small Reward analysis for completeness task (T5-large model): training steps vs. rewards (left), mean and variance of rewards (right).}
    % \label{fig:comp-reward-analysis}
\end{figure}

\section{Supplementary KL Divergence Analysis}
\FloatBarrier
\label{appendix:klD}
\begin{figure}[H]
    \centering
    \includegraphics[width=0.95\linewidth]{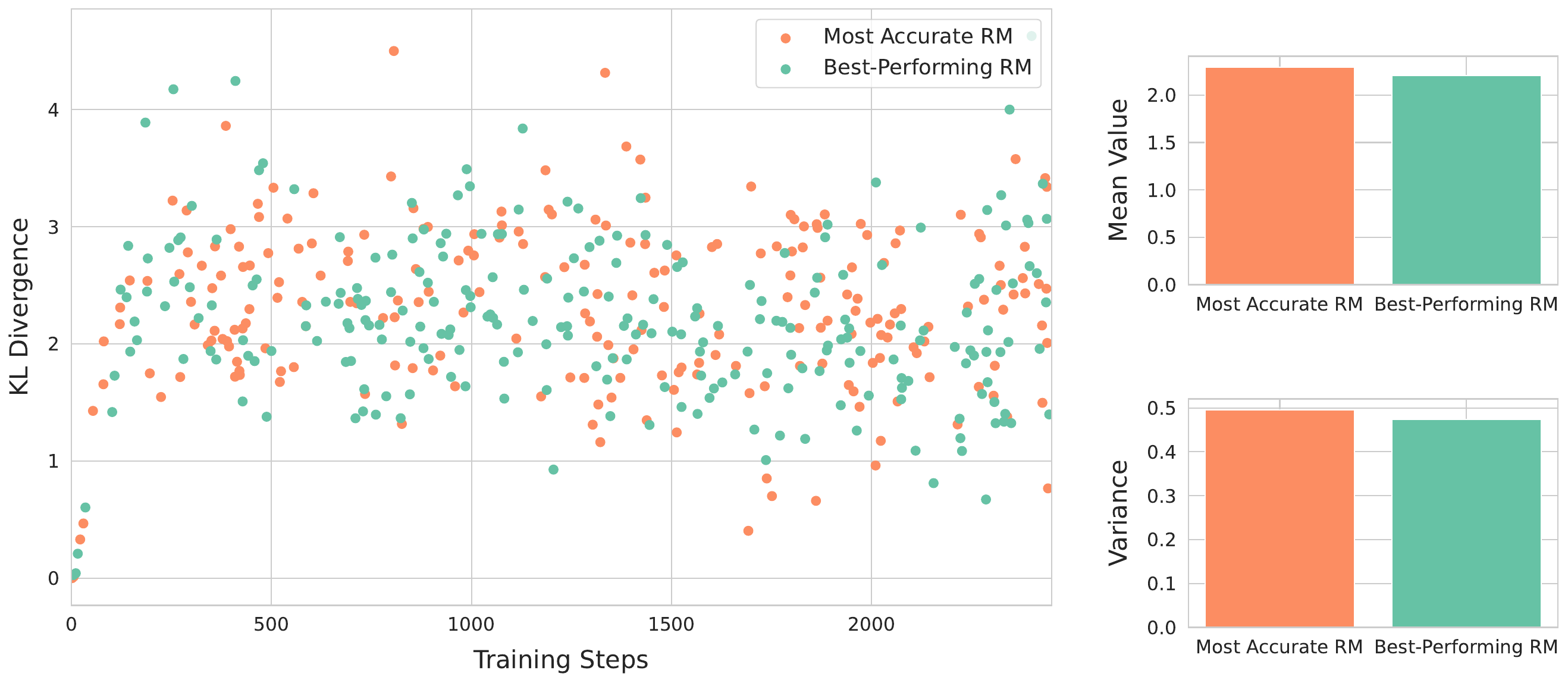}
    \caption{\small Relevance task KL divergence (T5-base model): training steps vs. KL divergence (left), mean and variance (right).}
    % \label{fig:rel-reward-analysis-3}
\end{figure}

\begin{figure}[H]
    \centering
    \includegraphics[width=0.95\linewidth]{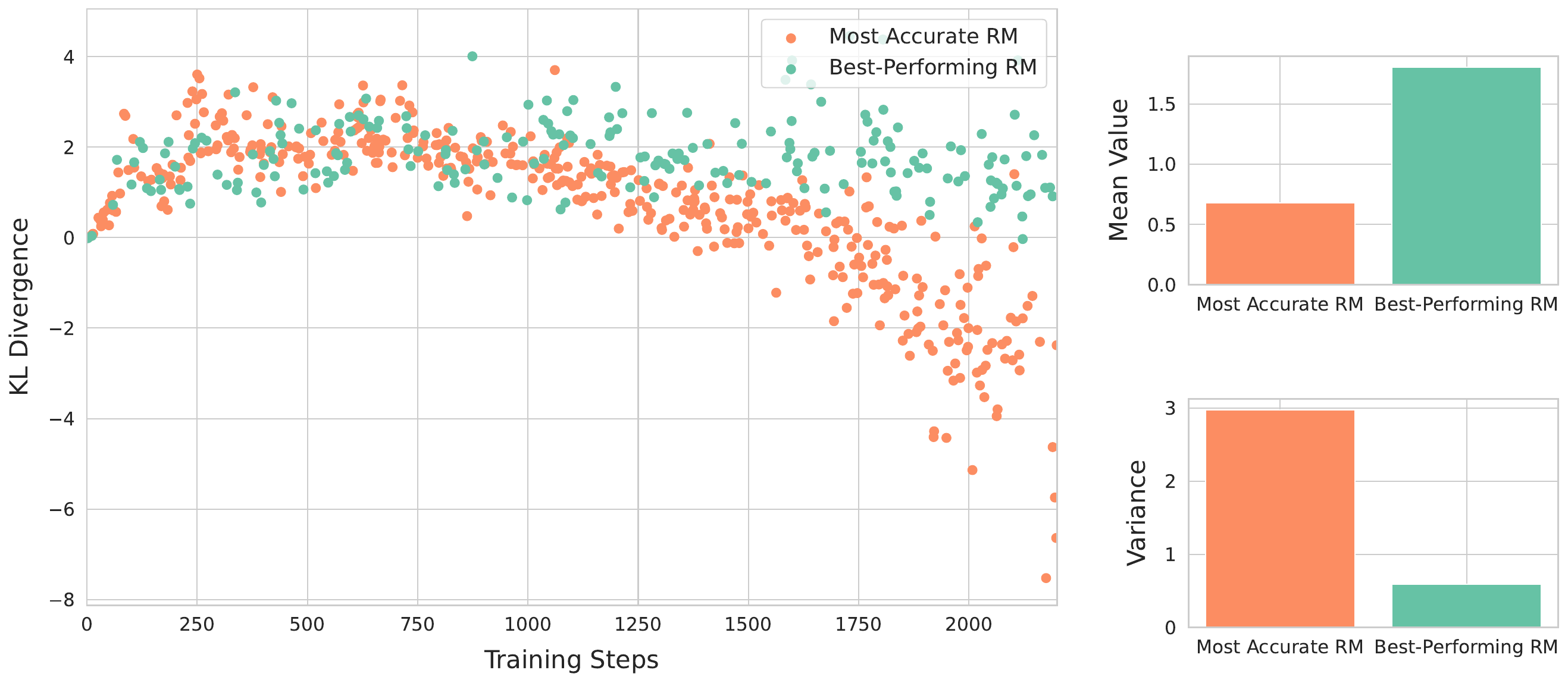}
    \caption{\small Factuality task KL divergence (T5-base model): training steps vs. KL divergence (left), mean and variance (right).}
    % \label{fig:fact-reward-analysis-3}
\end{figure}

\begin{figure}[H]
    \centering
    \includegraphics[width=0.95\linewidth]{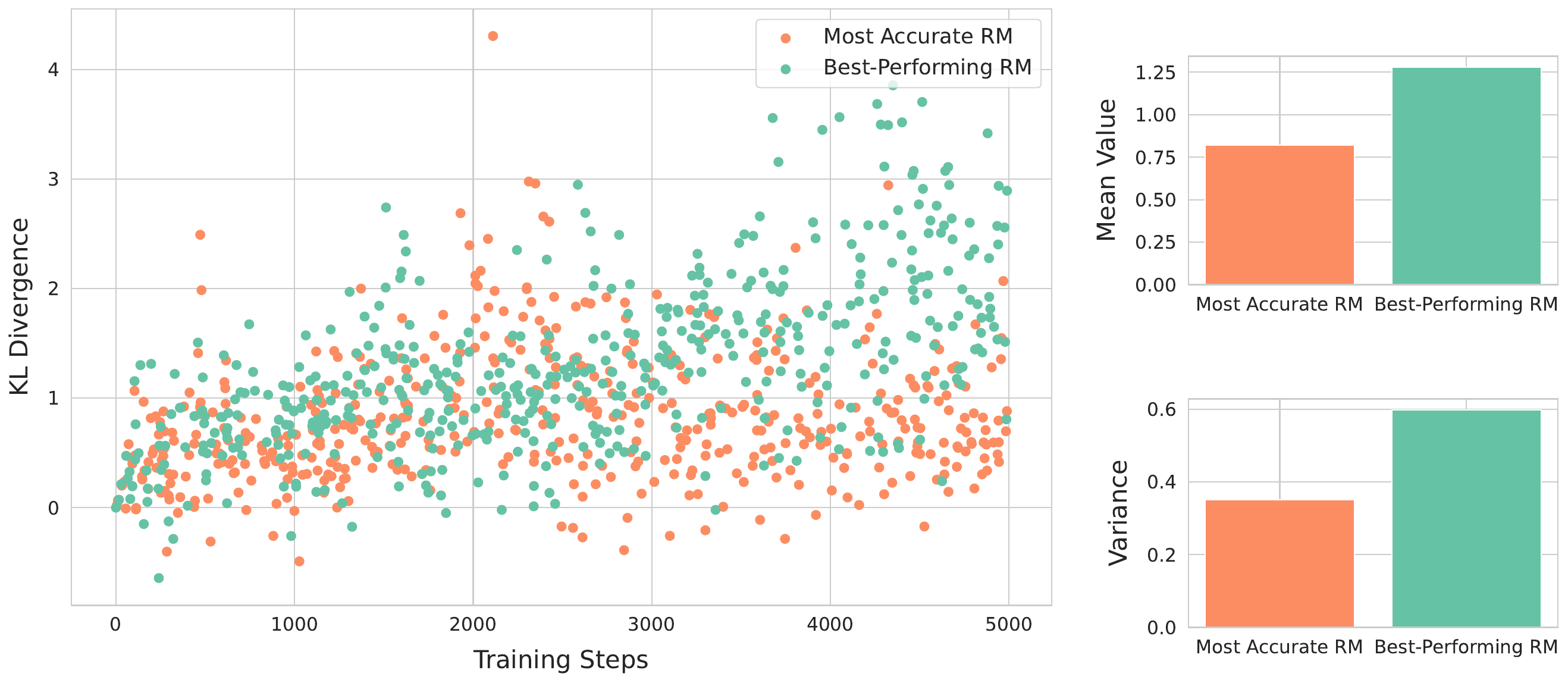}
    \caption{\small Completeness task KL divergence (T5-base model): training steps vs. KL divergence (left), mean and variance (right).}
    % \label{fig:comp-reward-analysis-3}
\end{figure}

\begin{figure}[H]
    \centering
    \includegraphics[width=0.95\linewidth]{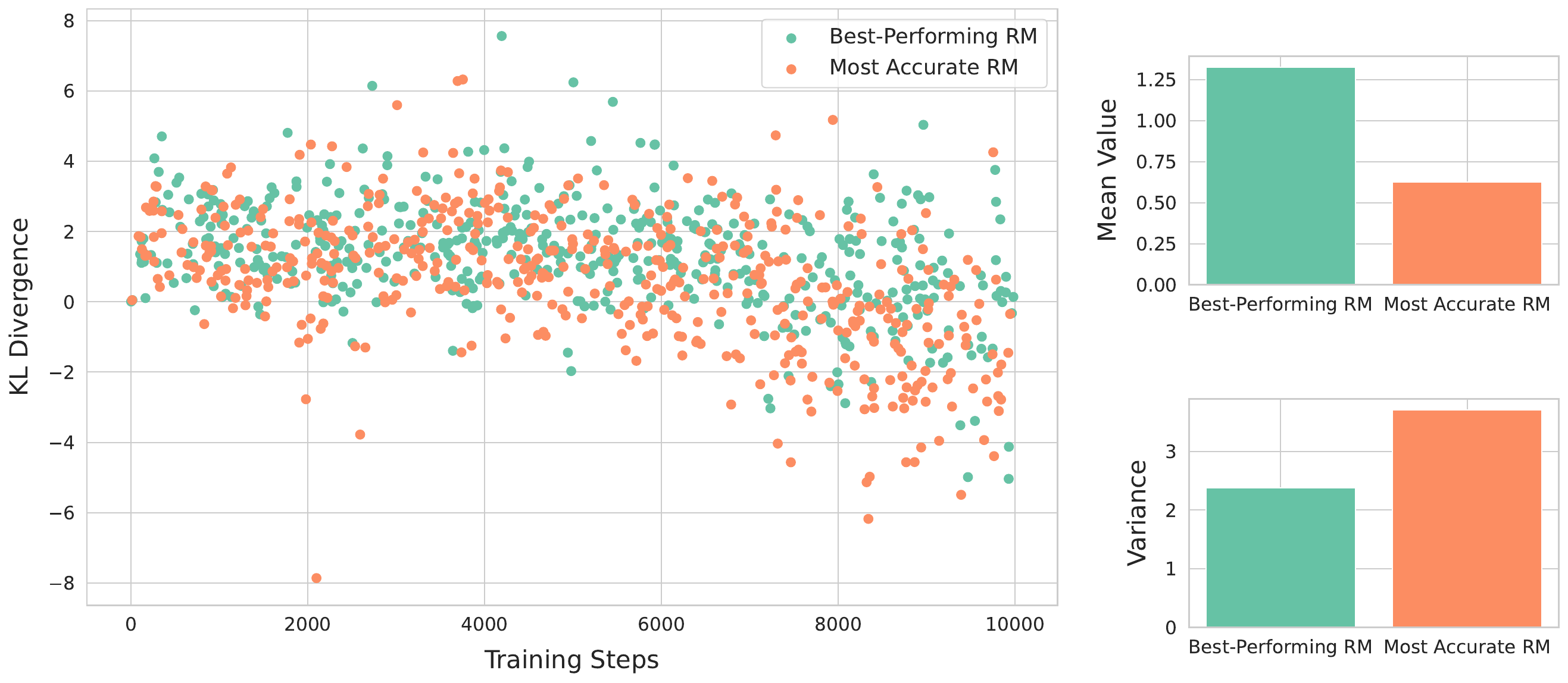}
    \caption{\small Relevance task KL divergence (T5-large model): training steps vs. KL divergence (left), mean and variance (right).}
    % \label{fig:rel-reward-analysis-3}
\end{figure}

\begin{figure}[H]
    \centering
    \includegraphics[width=0.95\linewidth]{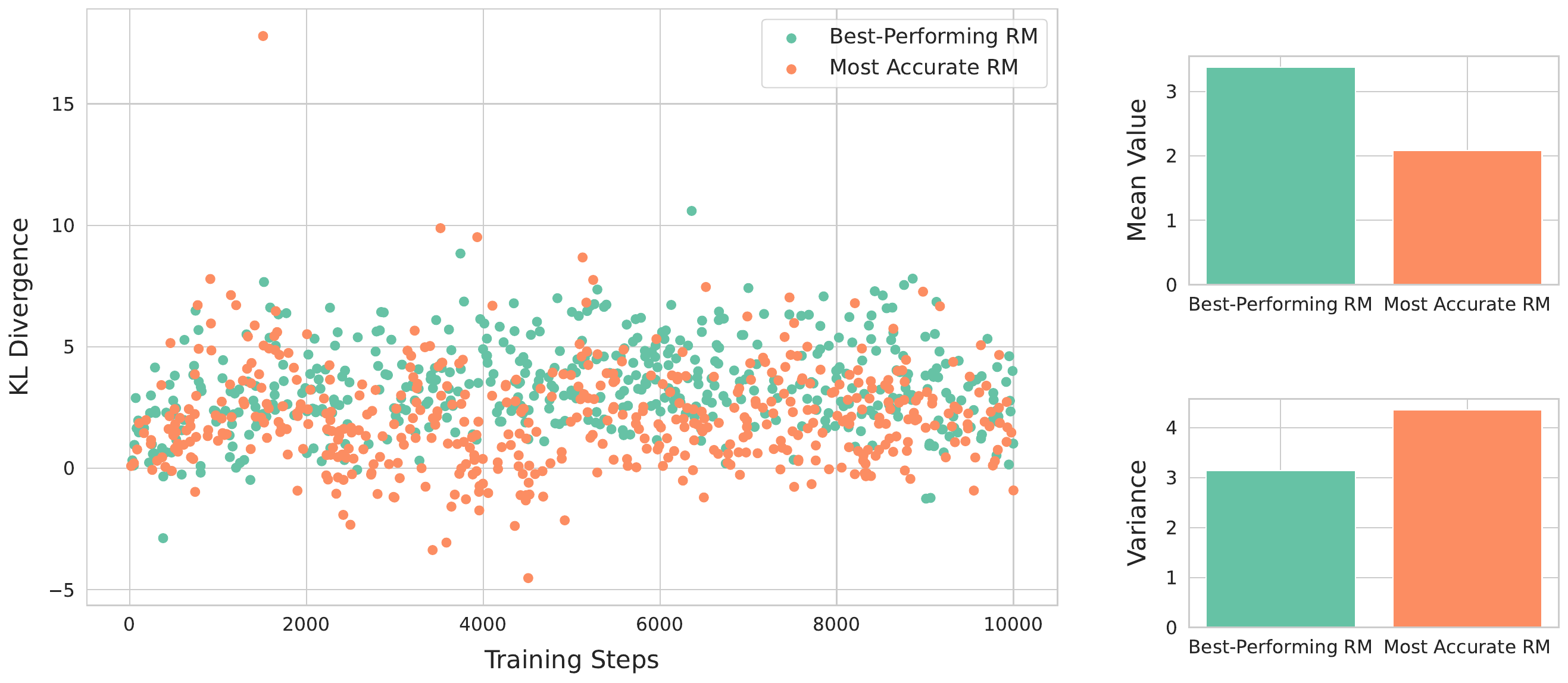}
    \caption{\small Factuality task KL divergence (T5-large model): training steps vs. KL divergence (left), mean and variance (right).}
    % \label{fig:fact-reward-analysis-3}
\end{figure}

\begin{figure}[H]
    \centering
    \includegraphics[width=0.95\linewidth]{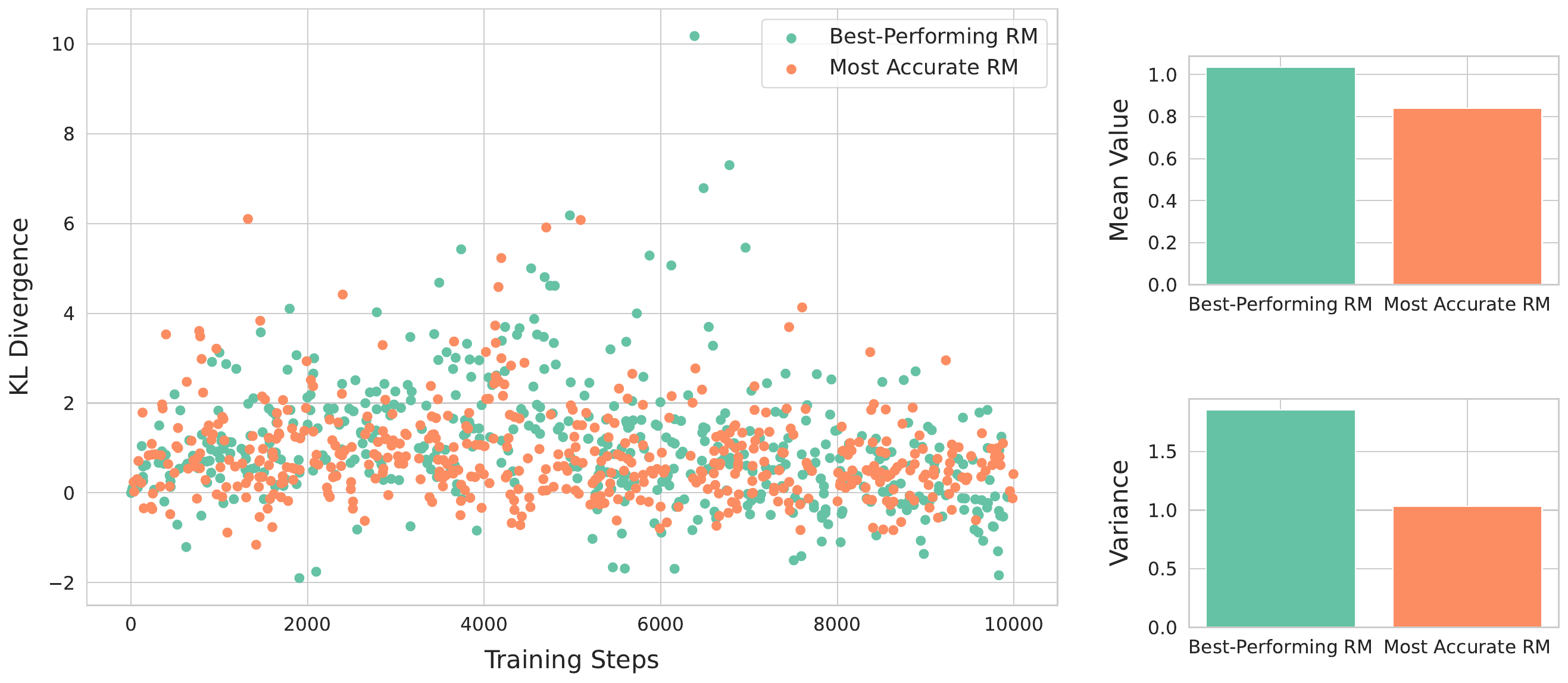}
    \caption{\small Completeness task KL divergence (T5-large model): training steps vs. KL divergence (left), mean and variance (right).}
    % \label{fig:comp-reward-analysis-3}
\end{figure}

\section{Hyperparameter Settings}
\label{appendix:hyperparameters}
\FloatBarrier
\begin{table}[H]
\footnotesize
\centering
\scalebox{1}{%
\begin{tabular}{@{}lc@{}}
\toprule[1.0pt]
\textbf{Model Component}       & \textbf{Setting}              \\ \midrule[1.0pt]
Input Padding Side    & Right                         \\ 
Top-K Sampling        & 20                            \\ 
Temperature           & 0.7                           \\ 
Value Model           & T5-base                       \\ 
Freeze Value Model    & False                         \\ 
Policy-Value Sharing  & False                         \\ \bottomrule[1.0pt]
\end{tabular}
}
\caption{\small Model Configuration Hyperparameters}
\label{tab:model_config}
\end{table}

\begin{table}[H]
\footnotesize
\centering
\scalebox{1}{%
\begin{tabular}{@{}lc@{}}
\toprule[1.0pt]
\textbf{Reward Model}             & \textbf{Setting}             \\ \midrule[1.0pt]
Relevance Model (Positive Reward)   & 0.3                \\ 
Relevance Model (Negative Reward)   & -0.3               \\ 
Factuality Model (Positive Reward)  & 0.5                \\ 
Factuality Model (Negative Reward)  & -0.5               \\ 
Completeness Model (Mean)           & -0.4468            \\ 
Completeness Model (Std)            & 8.3012             \\ 
Completeness Model (Bias)           & 0.0                \\ 
Completeness Model (Scale)          & 0.3                \\ \bottomrule[1.0pt]
\end{tabular}
}
\caption{\small Reward Model Hyperparameters}
\label{tab:reward_model}
\end{table}

\begin{table}[H]
\footnotesize
\centering
\scalebox{1}{%
\begin{tabular}{@{}lc@{}}
\toprule[1.0pt]
\textbf{Environment Parameter}   & \textbf{Setting}              \\ \midrule[1.0pt]
Maximum Input Length    & 1024                          \\ 
Maximum Generated Length & 200                           \\ 
Train Samples per Input & 4                             \\ \bottomrule[1.0pt]
\end{tabular}
}
\caption{\small Environment Configuration Hyperparameters}
\label{tab:env_config}
\end{table}

\begin{table}[H]
\footnotesize
\centering
\scalebox{1}{%
\begin{tabular}{@{}lc@{}}
\toprule[1.0pt]
\textbf{PPO Parameter}           & \textbf{Setting}               \\ \midrule[1.0pt]
KL Coefficient          & 0.3                            \\ 
Lambda                  & 0.95                           \\ 
Gamma                   & 1.0                            \\ 
Policy Gradient Coef.   & 1.0                            \\ 
Value Function Coef.    & 1.0                            \\ 
Clip Range (Policy)     & 0.2                            \\ 
Clip Range (Value)      & 0.2                            \\ 
Whiten Rewards          & True                           \\ \bottomrule[1.0pt]
\end{tabular}
}
\caption{\small PPO Training Hyperparameters}
\label{tab:ppo_hyperparams}
\end{table}

\begin{table}[H]
\footnotesize
\centering
\scalebox{1}{%
\begin{tabular}{@{}lc@{}}
\toprule[1.0pt]
\textbf{Training Parameter}        & \textbf{Setting}             \\ \midrule[1.0pt]
Total Episodes            & 80,000                       \\ 
Learning Rate             & 0.00001                      \\ 
Warmup Steps              & 100                          \\ 
PPO Epochs per Rollout    & 4                            \\ 
KL Threshold              & 20.0                         \\ 
Clip Gradients            & False                        \\ 
Max Gradient Norm         & 0.5                          \\ 
Random Seed               & 42                           \\ \bottomrule[1.0pt]
\end{tabular}
}
\caption{\small Training Procedure Hyperparameters}
\label{tab:train_proc}
\end{table}

\end{document}